%% file: aaai_main.tex
\documentclass[letterpaper]{article} 
\usepackage{aaai2026}  
\usepackage{times}  
\usepackage{helvet}  
\usepackage{courier}  
\usepackage[hyphens]{url}  
\usepackage{graphicx} 
\usepackage{natbib}  
\usepackage{caption} 
\usepackage{algorithm}
\usepackage{algorithmic}
\usepackage{graphicx}
\usepackage{amssymb}
\usepackage{amsmath}
\usepackage{enumitem}
\usepackage{booktabs}
\usepackage{multirow}
\usepackage{pifont}
\usepackage[table,xcdraw,dvipsnames]{xcolor}
\usepackage{subcaption}
\usepackage{newfloat}
\usepackage{listings}
\DeclareCaptionStyle{ruled}{labelfont=normalfont,labelsep=colon,strut=off} 
\floatstyle{ruled}
\newfloat{listing}{tb}{lst}{}
\floatname{listing}{Listing}
\nocopyright 

\title{$A^3$: Attention-Aware Accurate KV Cache Fusion for Fast \\ Large Language Model Serving}
\author{
    Yuechi Zhou\textsuperscript{\rm 1,2}, 
    Yi Su\textsuperscript{\rm 1}, 
    Jianxin Zhang\textsuperscript{\rm 1}, 
    Juntao Li\textsuperscript{\rm 1}\thanks{Corresponding author.}, 
    \\Qingrong Xia\textsuperscript{\rm 2}, 
    Zhefeng Wang\textsuperscript{\rm 2}, 
    Xinyu Duan\textsuperscript{\rm 2}, 
    Baoxing Huai\textsuperscript{\rm 2}
}
\affiliations{
    \textsuperscript{\rm 1}School of Computer Science and Technology, Soochow University \textsuperscript{\rm 2}Huawei Cloud

    yczhouyc@stu.suda.edu.cn
}

\usepackage{bibentry}

\begin{document}

\maketitle

\begin{abstract}
Large language models (LLMs) have demonstrated strong capabilities in processing long contexts, enabling them to tackle tasks involving long textual inputs such as multi-turn conversations, legal documents, or retrieved documents in Retrieval-Augmented Generation (RAG) systems. 
However, despite their ability to handle long sequences, the resulting decoding latency and memory overhead remain substantial, posing challenges for real-world deployment.
Recent advances in KV Cache reuse have shown potential to mitigate these costs, but still suffer from notable performance degradation. To address this issue, we conduct an in-depth investigation of recomputation-based reuse methods and observe that the recomputed tokens often fail to align with the context segments most relevant to the question. This misalignment hinders proper updates to the critical contextual representations. 
Therefore, we propose the \textbf{A}ttention-\textbf{A}ware \textbf{A}ccurate KV Cache Fusion algorithm ($A^3$), which precomputes and selectively fuses the KV Cache of text chunks based on their relevance to the question, achieving accurate integration with minimal computational overhead.
Extensive experiments on various benchmarks and LLMs demonstrate that $A^3$ achieves the best task performance compared to four baselines while reducing the time-to-first-token (TTFT) by 2×.

\end{abstract}

\input{sections/Introduction}

\input{sections/Related_Work}
\input{sections/Method}

\input{sections/Experiments}
\input{sections/Conclusion}

\bibliography{custom}

\newpage
\input{sections/supplementary.tex}
\end{document}

%% file: sections/Introduction.tex
\section{Introduction}
\label{sec:introduction}
Large language models (LLMs) \cite{qwen,zhao2024surveylargelanguagemodels,guo2025deepseek,llama3} have demonstrated remarkable capabilities in processing long-context inputs across a wide range of tasks, such as multi-turn dialogue generation \cite{yi2024survey}, long-document question answering \cite{qasper}, or evidence aggregation in RAG systems \cite{wang2024searching,li2024retrieval}.

Nevertheless, processing long textual inputs introduces substantial time and memory costs \cite{wu2024loongserve, zhong2024distserve} during autoregressive decoding, which degrades the user experience.
To accelerate inference, researchers adopt techniques such as model pruning \cite{LLM_Pruner, SparseGPT}, model quantization \cite{GPTQ, AWQ}, and token eviction \cite{H20, chen2024sepllm}. Recently, a promising solution is cross-request KV Cache reuse, which precomputes the KV Cache of potentially repeated text chunks across different inputs and reuses them in the future to reduce KV Cache computation.

\begin{figure}[t]
  \centering
  \includegraphics[trim=5 5 0 5, width=0.45\textwidth]{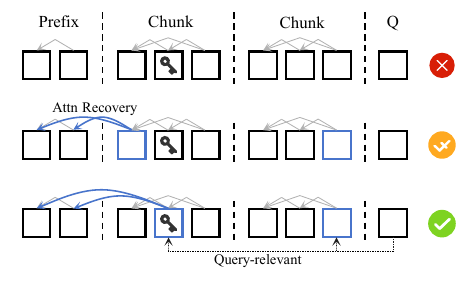}
  \caption{Blue blocks means the corresponding KVs are recomputed. (Top) Without recomputation. (Middle) Recomputation applied but key information missed. (Bottom) Key information updated based on query-aware attention.}
  \label{fig: intro pic}
\end{figure}

However, KV Cache reuse is lossy, with the loss primarily arising from the mismatch between the encoded positions during precomputation and the actual positions during inference (position loss), as well as the attention loss where later text blocks miss the attention from earlier ones. \citet{pcw, helet} utilize the properties of ROPE to achieve position recovery, while \citet{yao2025cacheblend, hu2024epic} use recomputation-based methods to update the KV Cache for important tokens to approximate attention recovery. Nevertheless, these methods still exhibit a significant performance gap compared to not using any reuse methods, particularly in long-text key information retrieval tasks.

We believe that position recovery is indispensable, but further recomputation yields limited improvements. We find that whether it is \citet{yao2025cacheblend} updating tokens according to the largest value difference or \citet{hu2024epic} using the sink rule \cite{Streamingllm} to update the head and tail tokens of chunks, both approaches neglect the role of the user's query. This leads to a misalignment between the recomputed tokens and the tokens focused on by the user query, which is important in retrieval tasks. High-attention tokens are not involved in the recomputation, while the updated tokens may be irrelevant to the query.

Therefore, we propose Attention-Aware Accurate KV Cache Fusion, which selects recomputation tokens based on the attention from the user query to the documents, enabling accurate KV Cache reuse. Figure \ref{fig: intro pic} provides a brief visual illustration. Our experiments across three LLMs and three benchmarks demonstrate that $A^3$ achieves the best task performance among four baselines, while maintaining a comparable computational overhead. Furthermore, $A^3$ is orthogonal to KV Cache compression methods, such as the quantization method KIVI \cite{KIVI} and the eviction method SnapKV \cite{li2024snapkv}. Given that KV reuse only optimizes the time latency of the first token, we additionally combined the token eviction method to optimize TPOT as an acceleration extension for $A^3$. This yields greater throughput and reduced decoding time at a minimal cost.

In summary, our contributions are as follows:

\begin{itemize}
    \item We propose $A^3$, a novel Attention-Aware Accurate KV Cache reuse algorithm that selects recomputation tokens based on the attention from the user query to the document, effectively bridging the mismatch between updated tokens and query-relevant content.
    \item We conduct extensive experiments on three LLMs across three long-context benchmarks, showing that $A^3$ achieves the best performance among baselines while maintaining comparable computational overhead.
    \item We further design a lightweight acceleration extension by integrating token eviction, which significantly reduces the TPOT and improves inference throughput, making $A^3$ practical for real-world deployment.
\end{itemize}

%% file: sections/Related_Work.tex
\section{Related Work}
\label{sec:related work}

\subsection{KV Cache Reuse}


Work on KV Cache reuse primarily falls into two categories based on the position of the reused chunks. The first is prefix caching \cite{zheng2023efficiently, jin2024ragcache, liu2023cachegen}, which reuses only the prefix text chunk. In contrast, in this paper, we focus on reusing text chunks from arbitrary positions within the input. For example, \textit{Prompt Cache} \cite{gim2024promptcache} proposes a modular caching scheme based on Prompt Markup Language. APE \cite{ape} restores accuracy by incorporating a shared prefix and using tunable attention temperatures and scaling factors. Furthermore, PCW \cite{pcw} and PIE \cite{helet} try to solve position loss based on ROPE \cite{su2024roformer}. \citet{yao2025cacheblend} and \citet{hu2024epic} developed CacheBlend and LegoLink, employing recomputation techniques in an attempt to restore the cross-attention between text chunks. CacheBlend recomputes tokens where the concatenated value states exhibit significant deviation compared to true values. LegoLink, leveraging the properties of Attention Sink \cite{Streamingllm}, recomputes a few initial and final tokens within each text chunk. Our work follows the recomputation-based approach, aiming to identify a tuning-free, efficient, and accurate reuse paradigm.

\subsection{KV Cache Compression}
As the length of texts continues to increase, the memory overhead of KV Cache becomes increasingly non-negligible \cite{susinkq}. To address this issue, researchers have explored two main directions. One line of work focuses on compressing the KV Cache through quantization techniques, reducing high-precision representations to lower-precision ones. Representative methods include KIVI \cite{KIVI}, KVQuant \cite{hooper2024kvquant}, and GEAR \cite{GEAR}. Another line of work adopts token eviction strategies, which selectively retain the KV Cache of important tokens. For example, StreamingLLM \cite{Streamingllm} preserves the initial tokens in the prompt, H2O \cite{H20} retains tokens with higher accumulated attention scores, and SepLLM \cite{chen2024sepllm} prioritizes special tokens. Our method runs orthogonally with these approaches but has been rarely explored in practice. In this work, we attempt to integrate it with SnapKV \cite{li2024snapkv} as an extension of our proposed method.

%% file: sections/Method.tex
\section{Method}
\label{sec:method}

\subsection{Background}
In long-context scenarios, the input typically consists of three components: a system prompt $\mathcal{S}$, a question \textit{Q}, and a long text passage $D$ relevant to the question. These passages are often composed of multiple smaller segments, i.e., $D=\{D_1, D_2, \cdots,D_k\}$.
Given a transformer-based \cite{transformer} LLM with parameters $\theta$, the input $X=(\mathcal{S},D,Q)$ of length $n$, and the corresponding hidden state representation $\boldsymbol{\mathit{X}}_l \in \mathbb{R}^{b \times n \times h}$ in layer $l$, where $b$ is the batch size and $h$ represents the hidden size, the model first computes and caches the KV tensors during the prefill phase. Specifically, to eliminate redundant computation of key and value vectors, LLMs typically store the KV Cache of the input in advance, i.e., \(K_l=\boldsymbol{\mathit{X}}_lW^k_l\) and \(V_l=\boldsymbol{\mathit{X}}_lW^v_l\) are computed and cached, where \(W^k_l,W^v_l\in\mathbb{R}^{h\times h}\) are the weight matrices at layer $l$. After prefilling, LLM autoregressively decodes the next token with its KV Cache ($\in \mathbb{R}^{b \times 1 \times h}$), where the generated KV vectors are appended to the existing KV Cache before performing the attention computation.

\subsection{Attention-Aware Accurate Fusion}
Our method consists of three main stages: precomputation, position recovery, and attention recovery, along with an optional acceleration extension.


\subsubsection{Precomputation.}When reusing KV Cache, we first precompute and store the KV Cache of the system prompt $\mathcal{S}$ and reusable documents $D$, which can be formulated as:
\begin{equation}
    K_i, V_i = \textrm{LLM}_\theta(i), \ i\in\{\mathcal{S}, D_1, D_2, \cdots,D_k\},
\end{equation}
where $\textrm{LLM}_\theta(\cdot)$ returns the KV Cache of text chunks for all layers. We concatenate each segment of $\{K,V\}_i\in\mathbb{R}^{b \times L \times |i| \times h}$ along the length dimension to obtain the $K_{cat}$ and $V_{cat}$, where $L$ represents the layer numbers of an LLM.
During precomputation, documents are precomputed for their corresponding KV Cache, and the KV tensors are stored in persistent storage (e.g., SSD).

\begin{figure*}[t]
  \centering
  \includegraphics[width=1\textwidth]{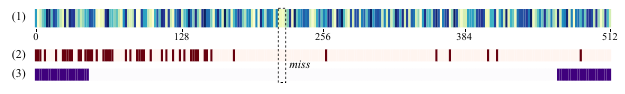}
  \caption{The mismatch between high-attention tokens and recomputed tokens. (1) The attention heatmap between the question and the document. We observe that both (2) the strategy of selecting tokens with the largest KV differences and (3) the strategy of selecting head and tail tokens of the document result in limited coverage of the high-attention tokens.}
  \label{fig: preliminary 1}
\end{figure*}

\begin{figure*}[t]
  \centering
  \includegraphics[width=1\textwidth]{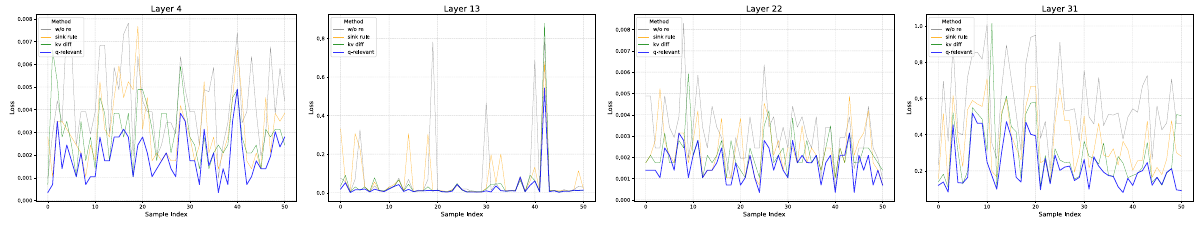}
  \caption{Attention recovery performance under different reuse strategies. The attention map obtained by recomputing question-relevant tokens is more consistent with the original attention map compared to other.}
  \label{fig: preliminary 2}
\end{figure*}

\subsubsection{Position Recovery.}Precomputing all chunks independently leads to incorrect positional encoding, since in the concatenated input to LLMs, a document’s position does not necessarily start from zero as it does during precomputation. For example, for document \( D_2 \) in input \( X=(\mathcal{S},\{D_1, D_2, \cdots,D_k\},Q) \), its position during precomputation ranges from 0 to \( |D_2|-1 \), while its position in \( X \) starts from \( |\mathcal{S}| + |D_1| \). Here, we leverage the properties of Rotary Position Embedding (RoPE) \cite{su2024roformer}, which is widely used in most LLMs, to achieve position recovery.

For the embedding vector $\boldsymbol{x}_m\in\mathbb{R}^h$ at position \(m\) in input \(X\) and given $\Theta \in \bigl\{\theta_i = 10000^{\frac{-2i}{h}}, i \in [0,1,\dotsc,\frac{h-1}{2}]\bigr\}$, multiplying $\boldsymbol{x}_mW_{k}$ on the left by a rotation matrix $R_{\Theta,m}^{h}$ can incorporate the positional information, where
\begin{equation}
  \notag
  \resizebox{\columnwidth}{!}{$   
  \displaystyle
  R_{\Theta,m}^{h}
  =
  \begin{pmatrix}
    \cos m\theta_{0} & -\sin m\theta_{0} & \cdots & 0 & 0 \\
    \sin m\theta_{0} &  \cos m\theta_{0} & \cdots & 0 & 0 \\
    \vdots & \vdots & \ddots & \vdots & \vdots \\
    0 & 0 & \cdots & \cos m\theta_{\frac{h}{2}-1} & -\sin m\theta_{\frac{h}{2}-1} \\
    0 & 0 & \cdots & \sin m\theta_{\frac{h}{2}-1} &  \cos m\theta_{\frac{h}{2}-1}
  \end{pmatrix}
  $}.
\end{equation}

Therefore, if we store the KV Cache of documents without positional embedding during precomputation and apply a global positional embedding to the concatenated KV Cache, i.e., applying the rotation matrix $R_{\Theta,m}^{d}$ to each position based on the length from 0 to \( |X|-1 \), the positions can be recovered. The concatenated Key tensor after position recovery can be formulated as:
\begin{equation}
    K'_{cat}=R_{\Theta,m}^{h}K_{cat},\  m=\{0,1,2,\cdots\}.
\end{equation}

\input{tables/longbench_results}
\subsubsection{Attention Recovery.} Now, we have obtained the KV pair $\langle K'_{cat},\ V_{cat} \rangle$ with recovered positions, but the attention from the later chunks to the earlier ones is still missing. Based on the example from Qasper \cite{longbench} shown in Figure \ref{fig: preliminary 1}, we observe that \ding{172}\textbf{existing recomputation-based methods tend to miss high-attention tokens}, achieving hit rates of only 30\% and 37.5\%, respectively. This limitation leads to suboptimal performance in the attention recovery process. As illustrated in Figure 3, we plot the L2 loss between the recomputed attention maps and the ground truth with two recomputation strategies. Both methods marginally reduce the attention loss compared to the baseline without recomputation, but the improvement is not substantial. In contrast, \ding{173}\textbf{we observe a more consistent reduction in attention loss when high-attention tokens are explicitly selected for recomputation}, across shallow (layer 4), middle (layers 13 and 22), and deep (layer 31) layers of the LLaMA3-8B-Instruct \cite{llama3}, as evaluated on sampled data from the Qasper. These two observations support the rationale behind our proposed method, and its effectiveness on specific tasks will be detailed in the experiment section. In the following, we proceed to describe our approach formally.

For an $L$-layer decoder-only LLM, we index the layer number $l$ from 0 to $L-1$. The input $X$ is first passed through the embedding layer and decoder block 0, producing the actual query ($Q_{true}$), key ($K_{true}$), and value ($V_{true}$) representations at layer 1. At this point, we aim to identify the document tokens $\{d_1, d_2, ...\}$ that are semantically relevant to the question $Q = \{q_0, q_1, ...\}$, based on the shallow representations. We compute the attention score between each question token $q_j$ and the document token $d_i$, yielding a question-relevant score $s$ for $d_i$ and $Re$
for the recomputation tokens:
\begin{align}
\label{eq: re}
    &s(d_i)=\sum\nolimits_{j} \mathrm{Softmax}(\frac{q_jK^T_{true}}{\sqrt{h}})|_{d_i}, \notag \\
    &Re=top_p(s(d_i)),\ p=rn,
\end{align}
where $r$ is a hyperparameter that determines the recomputation ratio and $n$ is the input length. Therefore, the attention output of layer 1 is: 
\begin{equation}
    O_1=\mathrm{Softmax}(Q_{true}[Re, Q]K^T_{true}/\sqrt{h})\cdot V_{true}.
\end{equation}

Therefore, the states updated at layer $l$ ($> 1$) is given by:
\begin{align}
    &\boldsymbol{X}_l=H(O_{l-1}), Q_l=\boldsymbol{X}_lW_l^q, \notag \\
    &K'_{cat_l}[Re,Q]\leftarrow \boldsymbol{X}_lW^k_l, \notag \\
    &V_{cat_l}[Re, Q]\leftarrow \boldsymbol{X}_lW^v_l, \notag \\
    &O_l=\mathrm{Softmax}(\frac{Q_lK'_{cat_l}}{\sqrt{h}})\cdot V_{cat},
\end{align}
where $H(\cdot )$ denotes the composition of layer normalization and the feedforward network applied to the attention output.



\subsubsection{Extention.} Moreover, we investigate the integration of our method with orthogonal eviction strategies to further optimize the generation latency of non-initial tokens and reduce overall KV Cache usage. Inspired by \citet{Streamingllm}, we first retain the tokens corresponding to the system prompt $\mathcal{S}$ and the question $Q$. Since the KV Cache for $\mathcal{S}$ is lossless and typically small (usually $|\mathcal{S}| < 50$), we choose to preserve it entirely without imposing a fixed-size constraint (e.g., the first 4 tokens). Given a maximum KV Cache capacity $C$, we then adopt the strategy proposed in SnapKV \cite{li2024snapkv} to adaptively select the top $C - |\mathcal{S}| - |Q|$ tokens at each layer and each attention head for retention. After the first token is generated, a unified KV Cache eviction is performed.


%% file: tables/longbench_results.tex
\begin{table*}[t]

\centering
\setlength{\tabcolsep}{4.1pt} 

\begin{tabular}{l|cccccccccccc}
\specialrule{1pt}{0pt}{2pt}
\multirow{4}{*}{Methods}  & \multicolumn{2}{c}{Single-Doc QA} & \multicolumn{2}{c}{Multi-Doc QA}& \multicolumn{2}{c}{Summarization}& \multicolumn{3}{c}{Few-shot Learning}& \multicolumn{2}{c}{Others} & \multirow{4}{*}{Avg.} \\
\cmidrule(lr){2-3}\cmidrule(lr){4-5}\cmidrule(lr){6-7}\cmidrule(lr){8-10}\cmidrule(lr){11-12}
& \rotatebox[origin=c]{30}{Qasper} & \rotatebox[origin=c]{30}{MultiQ} & \rotatebox[origin=c]{30}{Hotpot} & \rotatebox[origin=c]{30}{2WikiM} & \rotatebox[origin=c]{30}{GovRep} & \rotatebox[origin=c]{30}{MultiN} & \rotatebox[origin=c]{30}{TREC} & \rotatebox[origin=c]{30}{TriviaQ} & \rotatebox[origin=c]{30}{SAMSum} & \rotatebox[origin=c]{30}{PCount} & \rotatebox[origin=c]{30}{LCC}  & \\

\arrayrulecolor{black}\midrule
\multicolumn{13}{c}{LLaMA3-8B-Instruct} \\
\arrayrulecolor{black!20}\midrule
Vanilla  & 40.52 & 45.65 & 48.81 & 33.51 & 33.11 & 25.59 & 68.33 & 89.45 & 39.82 & 12.06 & 57.41 & 44.93 \\
\arrayrulecolor{black!20}\midrule
\textit{FullReuse} &4.63	&5.27	&3.48	&1.71	&13.96	&7.48	&0.67	&1.55	&1.89	&4.25	&25.43 &6.39 \\
\textit{PIE}	&34.71	&30.07	&38.55	&30.29	&27.55	&23.52	&61.33	&88.30	&39.30	&5.67	&64.12 &40.31 \\
\textit{CacheBlend} &35.03	& 32.82	& 36.97	& 28.53	& 28.47	& 24.45	& 67.67	& 87.58	& 39.07	& 5.67	& 56.29	& 40.23 \\
\textit{LegoLink} & 34.53	& 30.79	& 36.15	& 30.46	& 27.49	& 23.96	& 67.00	& 87.20	& 39.16	& 7.00	& 56.52	& 40.02 \\
\arrayrulecolor{black!20}\midrule
\textit{\textbf{ours}}	&\textbf{39.07}	&\textbf{36.46}	&\textbf{42.17}	&\textbf{32.56}	&\textbf{28.88}	&\textbf{24.55}	&\textbf{68.67}	&\textbf{89.25}	&\textbf{39.33}	&\textbf{8.67}	&\textbf{65.39} &\textbf{43.18} \\
\arrayrulecolor{black}\midrule
\multicolumn{13}{c}{Mistral-7B-Instruct-v0.2} \\
\arrayrulecolor{black!20}\midrule

Vanilla  & 20.09 & 42.59 & 24.10 & 18.37 & 32.35 & 25.43 & 62.33 & 88.60 & 40.47 & 3.92 & 62.68 & 38.27\\
\arrayrulecolor{black!20}\midrule
\textit{FullReuse} &5.00	&5.42	&0.97	&0.39	&22.37	&21.20	&0.33	&2.36	&2.38	&1.83	&27.80	&8.19 \\
\textit{PIE} &20.26	&38.85	&24.14	&17.15	&31.73	&25.62	&59.33	&\textbf{76.95}	&39.86	&5.30	&62.34	&36.50 \\
\textit{CacheBlend} &\textbf{21.72}	&39.17	&25.19	&18.19	&32.24	&\textbf{25.74}	&59.67	&75.04	&40.16	&3.39	&61.82	&36.58 \\
\textit{LegoLink} &21.38	&\textbf{40.14}	&25.46	&17.13	&32.03	&25.72	&59.67	&75.82	&40.16	&\textbf{4.34}	&63.57	&36.86\\
\arrayrulecolor{black!20}\midrule
\textit{\textbf{ours}} &21.12	&39.31	&\textbf{25.52}	&\textbf{18.38}	&\textbf{32.41}	&25.28	&\textbf{59.67}	&76.44	&\textbf{40.25}	&3.15	&\textbf{65.22}	&\textbf{36.98} \\
\arrayrulecolor{black}\bottomrule
\end{tabular}

\caption{Performance comparison of our method with \textit{FullReuse}, \textit{PIE}, \textit{CacheBlend}, \textit{LegoLink}, and up-bound method Vanilla on LongBench for LLaMA3-8B-Inst and Mistral-7B-Inst-v0.2. The best results are highlighted in \textbf{bold}.}
\label{table: longbench results}
\end{table*}

%% file: sections/Experiments.tex
\section{Experiments}
\label{sec:experiments}

\subsection{Experiment Setup}
\label{subsec: Exp setup}

\noindent \textbf{Datasets.} We select the following three widely used long-text benchmarks that are commonly adopted in both industry and academia to evaluate advanced models:
\begin{itemize}[label=\textbullet, left=0pt, labelsep=3pt, topsep=2pt, itemsep=0pt, partopsep=0pt, parsep=0pt]
    \item \textbf{LongBench} \cite{longbench}: This benchmark comprehensively evaluates the long-text understanding capabilities of LLMs. We select eleven representative datasets from it, covering a range of tasks including single-document and multi-document question answering, summarization, few-shot learning, code completion, and more.
    \item \textbf{Needle-in-a-Haystack} \cite{liu2024lost}: This dataset evaluates the ability of LLMs to retrieve key information from long contexts. Performance on this dataset reflects how well the reuse methods can recover critical details; failure to identify key content indicates that the method may impair the model’s retrieval capabilities.
    \item \textbf{Ruler} \cite{hsieh2024ruler}: Ruler is designed to evaluate the ability of LLMs to locate and reason over relevant evidence from long unannotated contexts. It challenges models to retrieve and utilize supporting information that is not explicitly highlighted, thereby testing their capacity for multi-hop and aggregation tasks. 
\end{itemize} 

\input{tables/needle_results}
\input{tables/ruler_results}
\noindent \textbf{Models and Baselines.} We adopt LLaMA3-8B-Inst \cite{llama3}, Mistral-7B-Inst-v0.2 \cite{mistral}, and Qwen2.5-7B-Inst \cite{qwen} as backbone models. For fair comparison, we primarily consider five baseline methods, including the following: 
\begin{itemize}[label=\textbullet, left=0pt, labelsep=3pt, topsep=2pt, itemsep=0pt, partopsep=0pt, parsep=0pt]
    \item Vanilla: It does not employ any modifications, such as KV Cache reuse. Vanilla performs inference with FP16 precision (applied consistently across all methods) and serves as the upper bound for all following reuse approaches.
    \item \textit{FullReuse}: It represents the most basic reuse method, where the saved KV Caches corresponding to $\mathcal{S}$ and $\mathcal{D}$ are simply concatenated without any further modifications.
    \item \textit{PIE} \cite{helet}: Building upon \textit{FullReuse}, it encodes the correct positional information for the concatenated KV Cache without any additional recomputation.
    \item \textit{CacheBlend} \cite{yao2025cacheblend}: A recomputation-based method, selectively recomputing the KV Cache for a subset of tokens after position recovery. Specifically, it updates the tokens with the largest discrepancies between the concatenated values and the ground-truth values.
    \item \textit{LegoLink} \cite{hu2024epic}: A recomputation-based method that, following the findings of StreamingLLM \cite{Streamingllm}, selects several tokens at the beginning and end of each document for KV Cache recomputation.
\end{itemize}

\noindent \textbf{Details.}
We split the input into several chunks for all benchmarks with a chunk size of 512, which is consistent with \textit{CacheBlend}. For \textit{CacheBlend}, we balance the performance and computation time by setting the recomputation rate to the commonly used value of 0.15. Similarly, for \textit{LegoLink}, we recompute the first and last 20 tokens of each chunk. 

\begin{figure*}[t]  
  \centering  
  \includegraphics[width=1\textwidth]{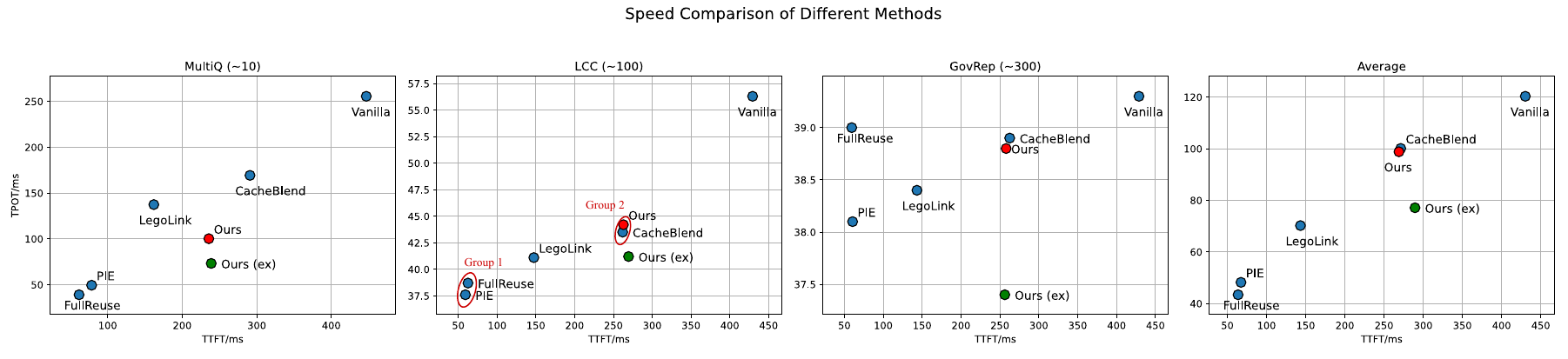}
  \caption{Comparison of inference performance among different reuse methods.}  
  \label{fig: inference performance}  
\end{figure*}

\input{tables/ablations}
\subsection{Task Performance}
Due to space constraints, several results on specific models are provided in the Appendix (Supplementary Material) for reference, such as LongBench results for Qwen2.5-7B-Inst. 
\subsubsection{Results on LongBench.}Table \ref{table: longbench results} presents results of various models and reuse methods on the LongBench. We report F1 scores for Single-Document QA, Multi-Document QA tasks, and TriviaQA; ROUGE-L \cite{chin2004rouge} for summarization tasks and SAMSum; Exact Match for TREC and PassageCount; and fuzzy string matching for evaluating code similarity in the code completion task (LCC).

Our method achieves \textbf{the best} average performance, particularly on LLaMA3-8B-Inst, where it consistently outperforms all other reuse baselines across every sub-dataset. Moreover, compared to the Vanilla setting, our approach achieves near lossless performance (36.98 vs 38.27 in Mistral). From Table \ref{table: longbench results}, we observe that position recovery is essential, as evidenced by the poor performance of \textit{FullReuse}. In addition, reuse methods perform well on summarization tasks such as MultiNews and SAMSum, likely because missing some key information has a limited impact on overall generation quality. In contrast, on QA tasks like Qasper and HotpotQA, baseline methods exhibit a noticeable drop in performance (40.52 $\to$ 35.03). While our method significantly outperforms them (39.07 $\to$ 35.03), indicating that our recomputation strategy effectively captures critical information necessary for producing correct answers.

\subsubsection{Results on Needle-in-a-Haystack.}The six subfigures in Figure \ref{fig: needle results} present results of six methods. We report the ROUGE score between the generated response and the inserted needle text. The x-axis represents the input text length in the current experiment, while the y-axis indicates the relative position at which the needle is inserted into the input. 

The observed results align well with intuition: position recovery is essential ($\textit{PIE}\gg\textit{FullReuse}$); attention recovery is also needed ($\textit{LegoLink} > \textit{PIE}$); the data-driven approach outperforms the rule-based one ($\textit{CacheBlend} > \textit{LegoLink}$); and the attention-guided recomputation method outperforms the one driven by KV Cache discrepancy ($\textit{Ours} > \textit{CacheBlend}$). Nevertheless, none of the reuse methods reach the upper bound (Vanilla $> all$). In summary, our method achieves \textbf{the best} performance in Ruler, which demonstrates the key information retrieval capability of $A^3$.

\subsubsection{Results on Ruler.}We report the Exact Match score for each sub-dataset, as shown in Table \ref{table: ruler results}. Our method achieves \textbf{the highest} average accuracy (84.17\%) among all KV reuse baselines on Qwen2.5-7B-Inst. Compared to \textit{FullReuse} (43.23\%) and \textit{PIE} (76.97\%), our method significantly improves robustness across both Single and Multi-key NIAH tasks. While \textit{CacheBlend} (83.88\%) and \textit{LegoLink} (78.88\%) perform well, our method consistently outperforms them, especially on challenging tasks like FWE and VT, highlighting its advantage in maintaining high accuracy under efficient KV reuse.

\subsection{Inference Performance}
\input{tables/ex}

During inference, we conduct experiments using the Transformers framework in combination with the FlashInfer \cite{yeflashinfer} acceleration module on NVIDIA 3090 GPUs. As shown in Figure \ref{fig: inference performance}, we evaluate the LLaMA3-8B-Inst model and report the TTFT and TPOT under different reuse strategies for generation lengths of 10, 100, and 300 tokens.

The six methods can be grouped based on their reuse and recomputation strategies. Group 1 includes \textit{PIE} and \textit{FullReuse}, which require no recomputation and therefore appear near the bottom-left corner of the plot. Group 2 includes $A^3$ and \textit{CacheBlend}, which recompute the same number of tokens, with the only difference being the overhead in KV difference computation and attention computation (Eq. \ref{eq: re}). \textit{LegoLink} lies between Group 1 and Group 2, representing a trade-off between reuse granularity and recomputation. Lastly, \textit{Vanilla}, which recomputes KV caches for the entire sequence, appears at the top-right of the figure due to its highest memory and computation cost.

Notably, our method \textbf{achieves efficiency comparable} to \textit{CacheBlend}, while outperforming it in task performance. In particular, our method yields a speedup of about 2× over the vanilla approach, and up to 3.3× acceleration can be observed on larger models (e.g., 70B) \cite{yao2025cacheblend}. We also separately measure the time cost of computing $Re$ (Eq. \ref{eq: re}), which takes only 0.1 ms on average for an input length of 6,000 tokens. This aligns with the observation in Figure \ref{fig: inference performance}, where our method and \textit{CacheBlend} appear close together.

When combined with eviction-based extension (\textit{ex}), we observe a shift toward the bottom-right in the plot (red dot $\to$ green dot), indicating a slight increase in TTFT (approximately 5 ms) in exchange for \textbf{improved further decoding efficiency}. For instance, in the MultiQA dataset, this leads to a 25\% reduction in TPOT. In our experiments, the maximum KV capacity $C$ is set to 1024, which corresponds to approximately 17\% of the input length. Additionally, we evaluate the maximum generation length and GPU memory savings before and after using \textit{ex}. It is important to note that the reuse methods themselves do not introduce additional GPU memory overhead. As shown in Figure \ref{fig:max_length}, when the maximum allowed GPU memory is 20GB, the maximum generation length achievable with \textit{ex} is comparable to that of the 24GB setting without \textit{ex} (with an input length of 7044 tokens). Furthermore, when the input length reaches 10K tokens, $A^3$ (\textit{ex}) reduces GPU memory consumption by approximately 4.4GB, which can then be repurposed for decoding. 

Furthermore, we compare the throughput of \textit{Vanilla}, the most efficient method \textit{FullReuse} (disregarding task performance), and our method with extension. The input length is 2,000 tokens, and the output length is 100 tokens. From the results shown in Figure~\ref{fig:throughput}, we can observe that the extension improves the throughput of $A^3$ and as the batch size increases, the extension leads to a significant improvement in throughput.

But what is the cost? We evaluate the impact of our extension strategy on three models across the LongBench and Ruler datasets, reporting results before and after applying KV eviction. A dash (``–'') indicates that the model fails to generate intelligible output. As shown in Table \ref{table: ex}, extension strategy does not lead to catastrophic performance degradation—despite being a combination of two lossy mechanisms. The overall performance \textbf{drop remains controllable}. For instance, on LongBench, the Mistral model experiences only a 1\% decrease (36.98 $\to$36.60) in average accuracy. Considering the inference speedup by \textit{ex} discussed earlier, this represents a practical trade-off between performance and efficiency. To the best of our knowledge, this is the first work to demonstrate the feasibility of jointly optimizing KV reuse and KV eviction.

We further analyze the impact of different recomputation ratios in Figure \ref{fig:ratio}, and provide additional experimental results and a code appendix in the supplementary material.

%% file: tables/needle_results.tex
\begin{figure*}[t]
  \centering
  \includegraphics[width=0.33\linewidth]{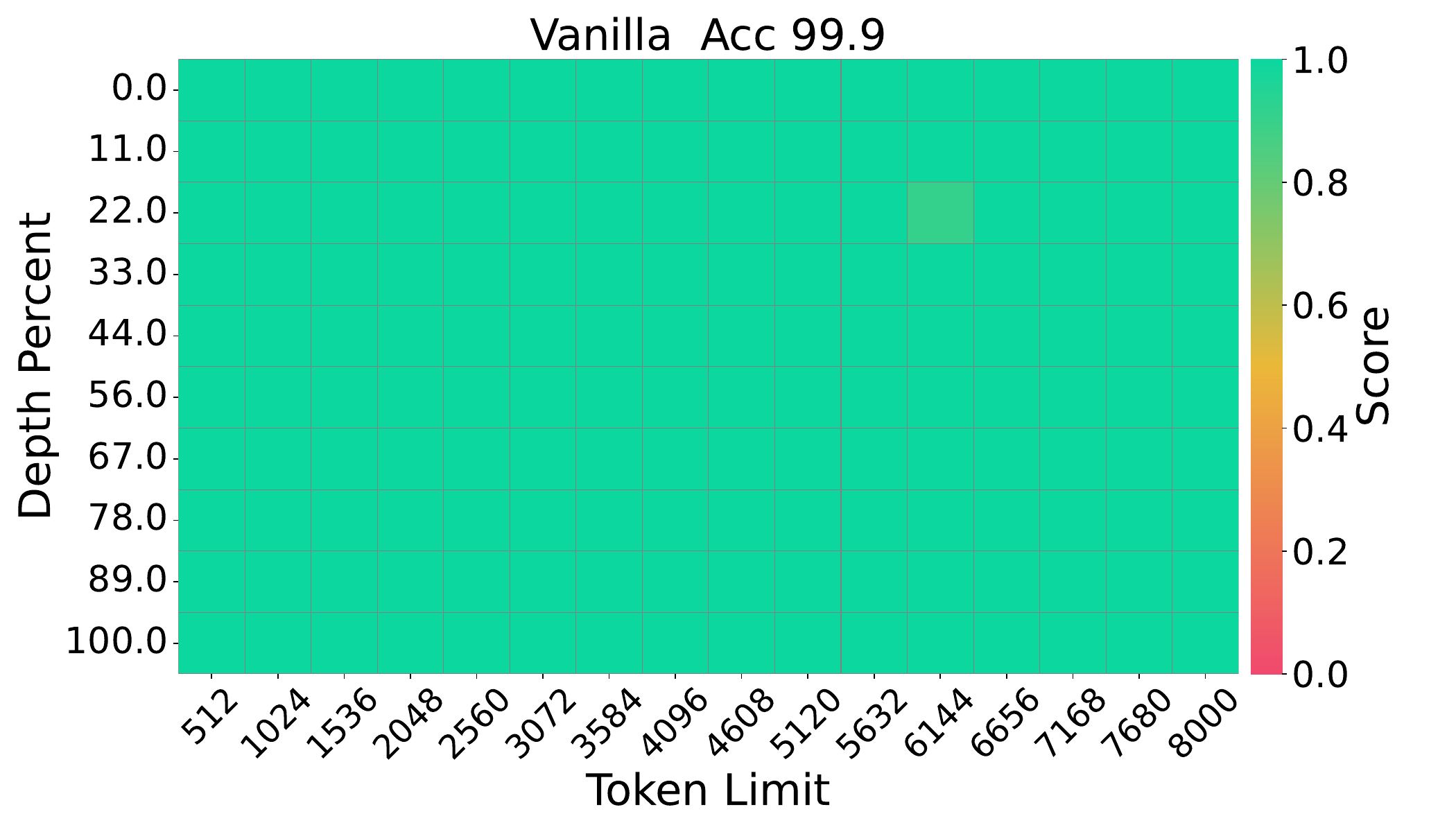}
  \includegraphics[width=0.33\linewidth]{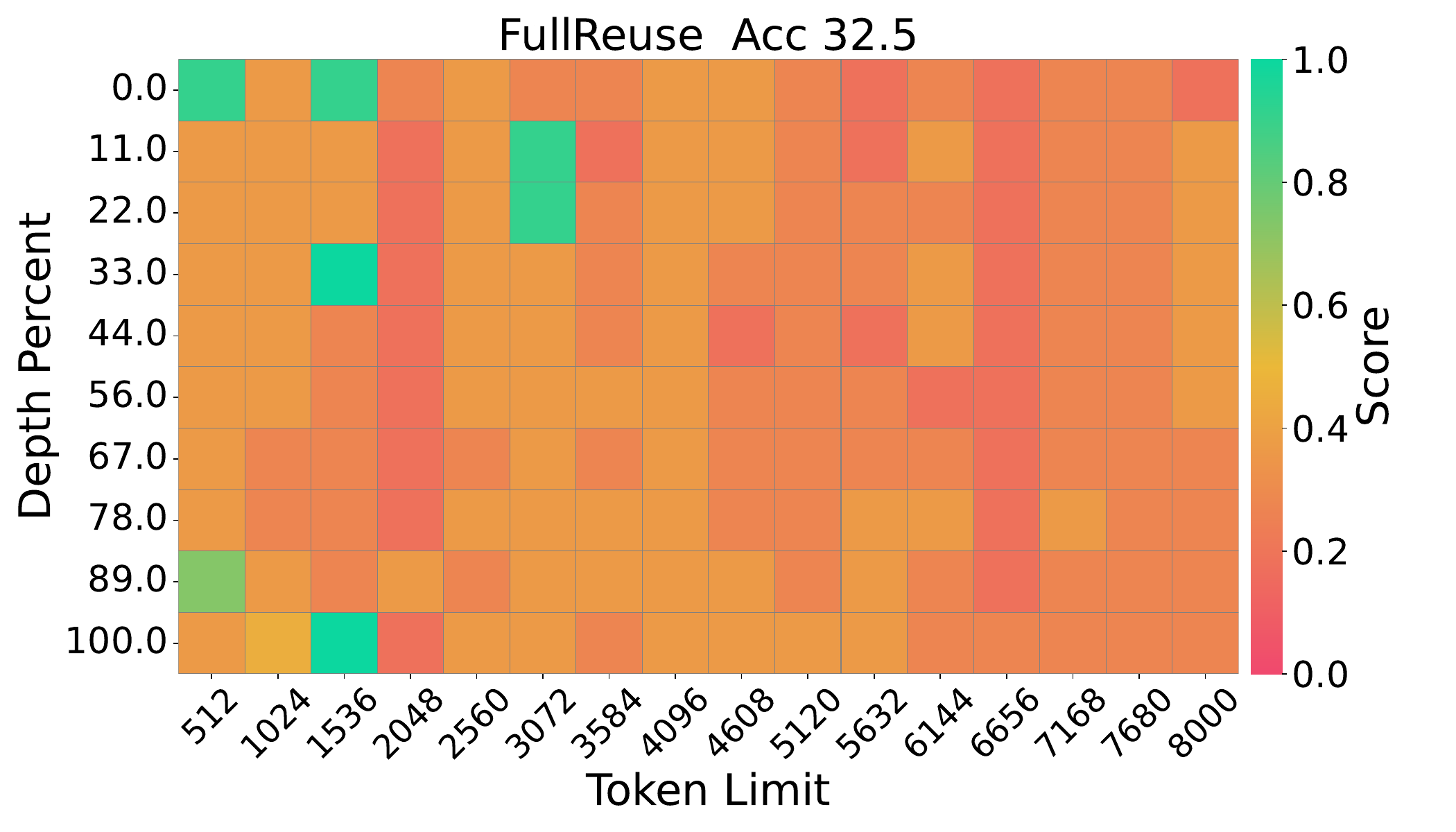}
  \includegraphics[width=0.33\linewidth]{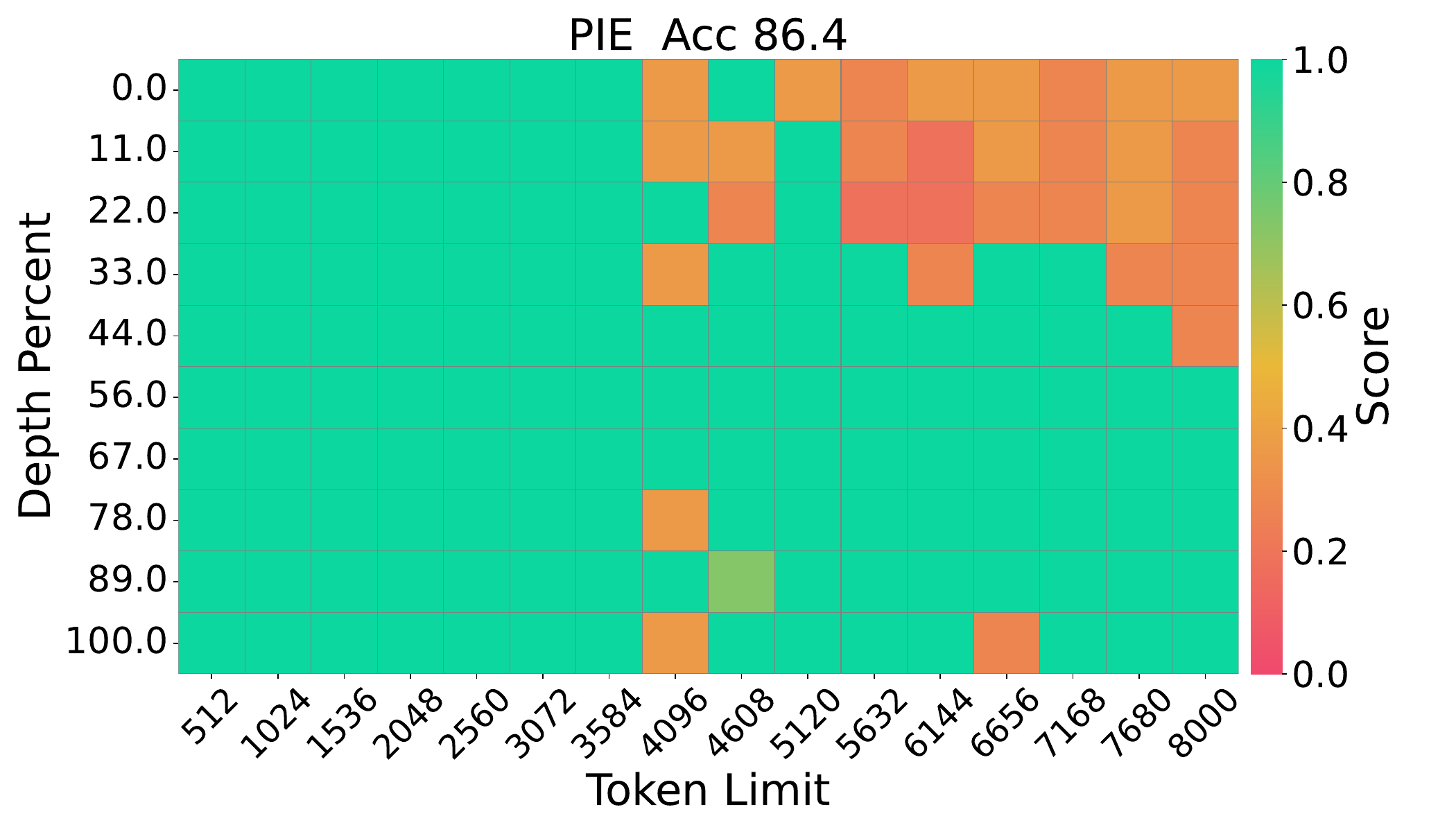} \\
  \includegraphics[width=0.33\linewidth]{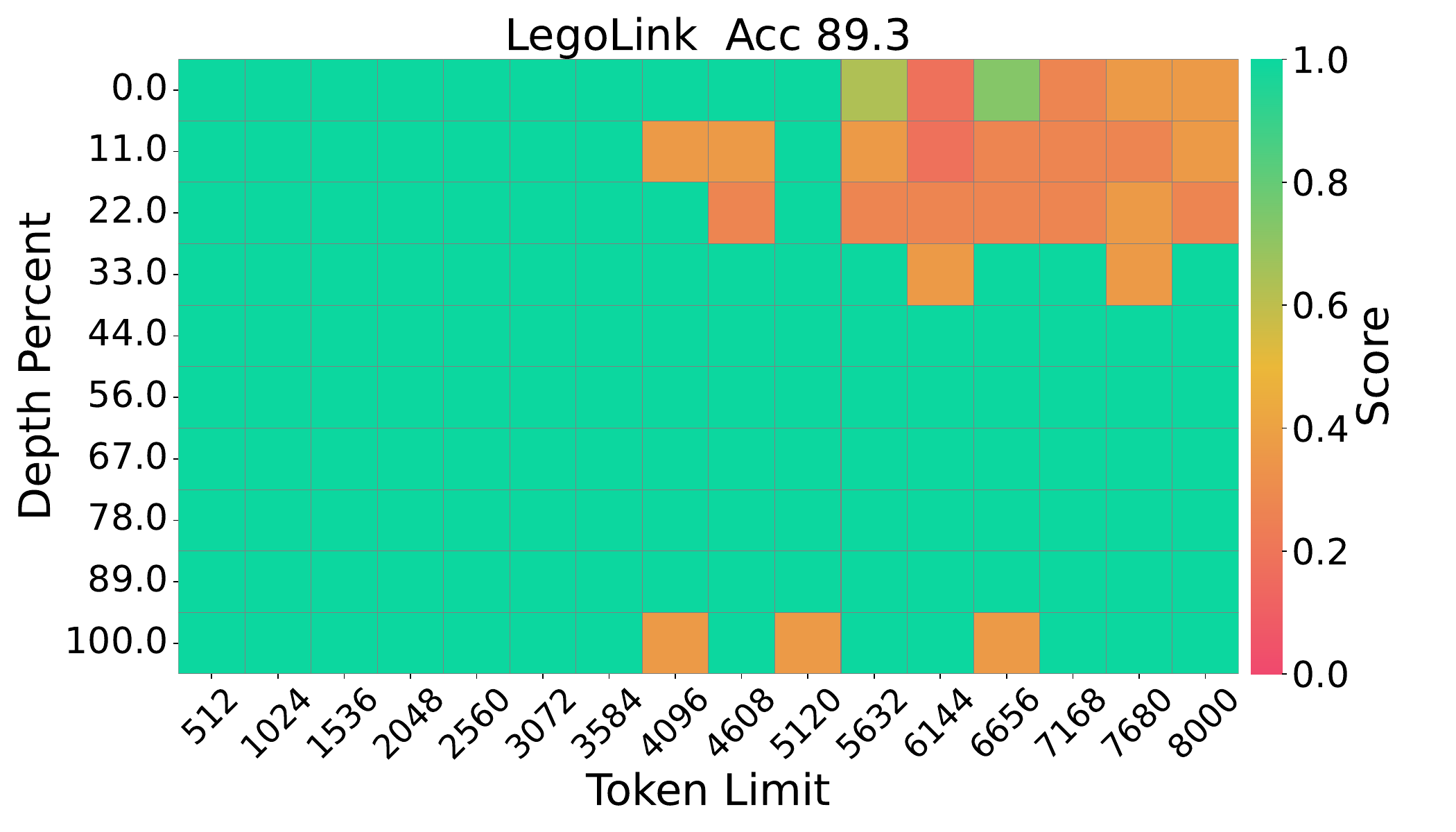}
  \includegraphics[width=0.33\linewidth]{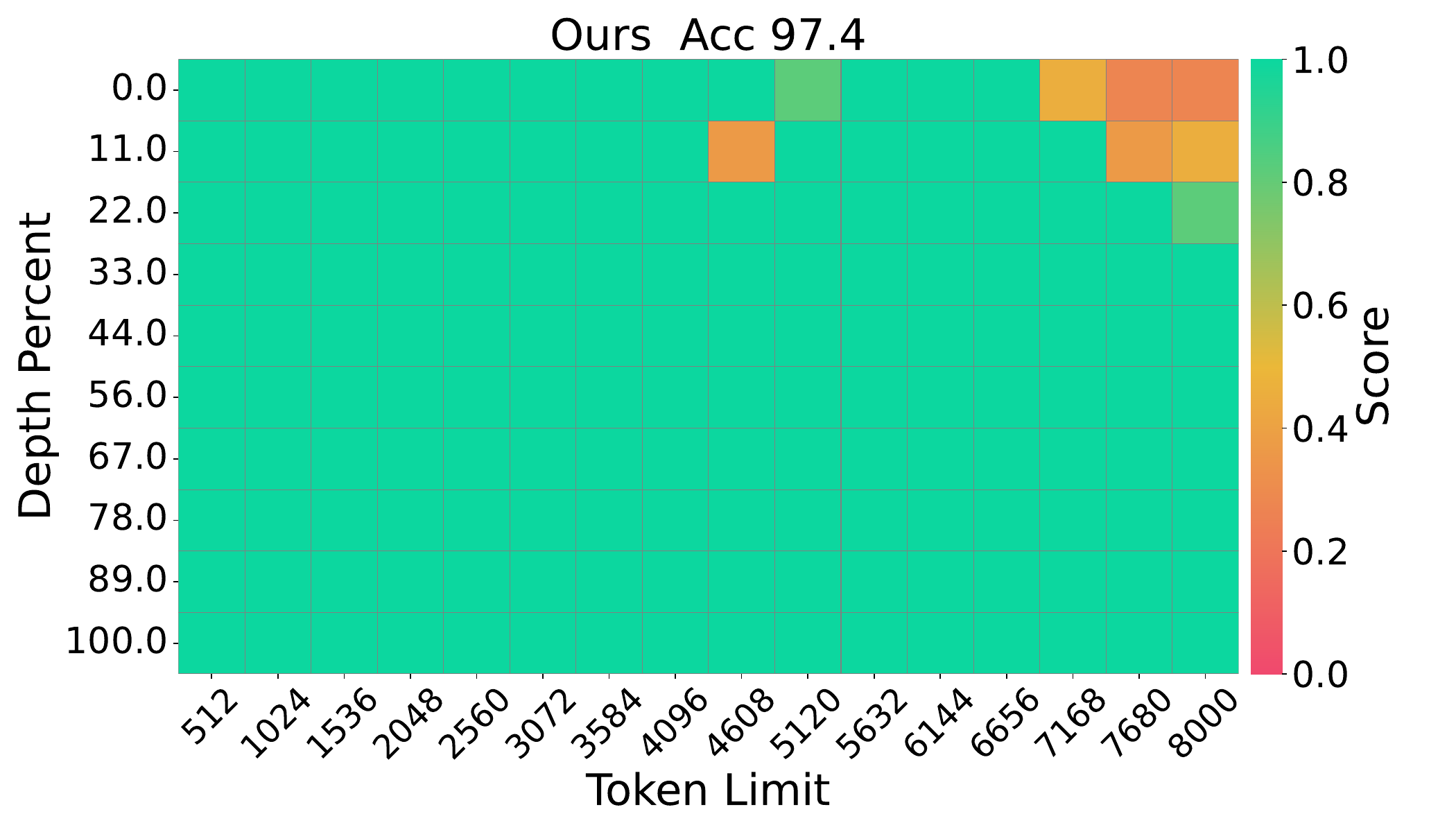}
  \includegraphics[width=0.33\linewidth]{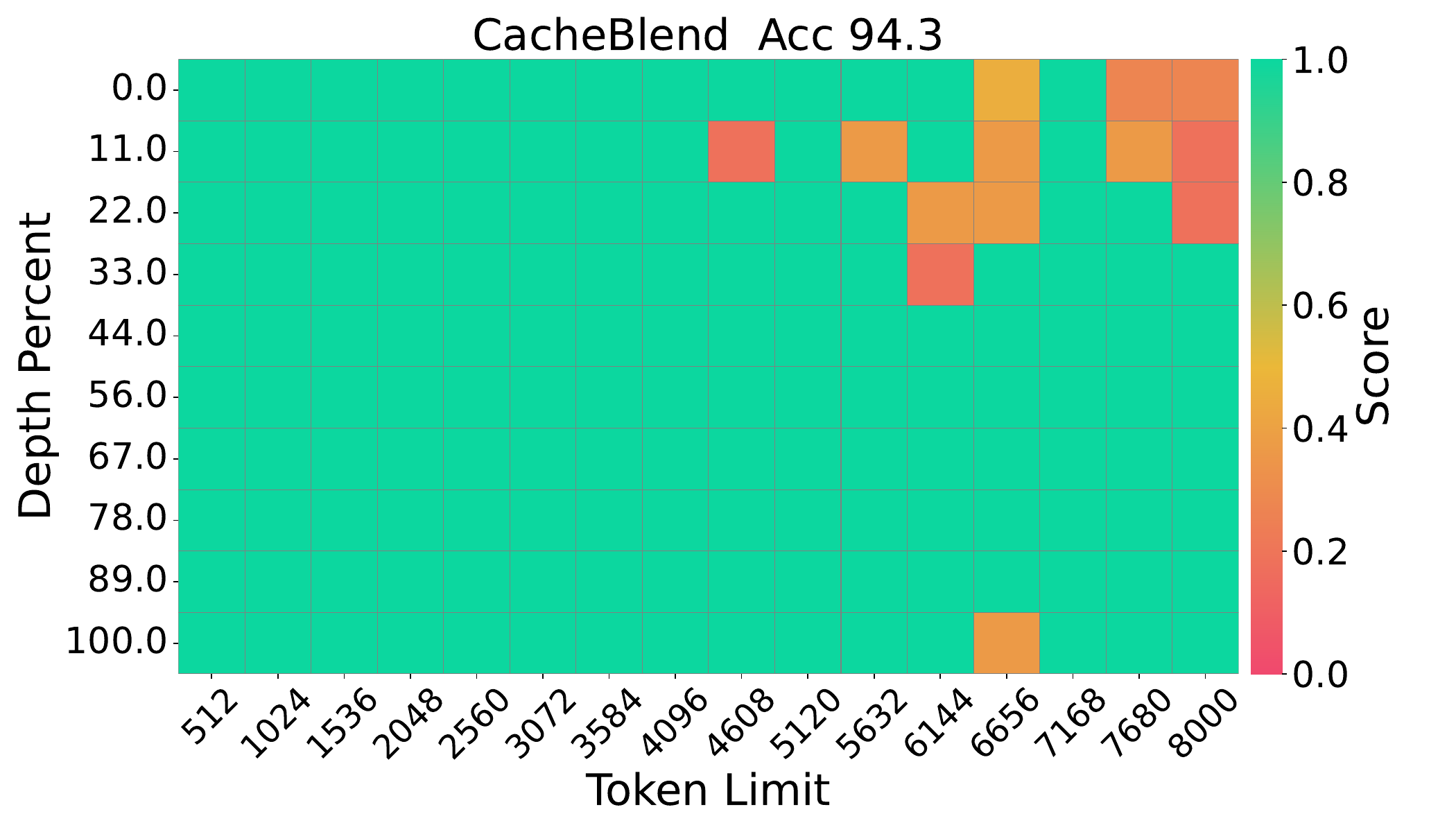}
  \caption{Results of Needle-in-a-Haystack on LLaMA-3-8B-Instruct. The vertical axis of the figure represents the depth percentage, and the horizontal axis represents the token length. Our method achieves the best retrieval performance.}
  \label{fig: needle results}
\end{figure*}

%% file: tables/ruler_results.tex
\begin{table*}[t]
\centering

\setlength{\tabcolsep}{4.2pt} 

\begin{tabular}{l|cccccccccccc}

\specialrule{1pt}{0pt}{2pt}
\multirow{4}{*}{Methods}  & \multicolumn{3}{c}{Single NIAH} & \multicolumn{3}{c}{Multi-key NIAH} & \multirow{4}{*}{\rotatebox[origin=c]{30}{MQuery}} & \multirow{4}{*}{\rotatebox[origin=c]{30}{MValue}} & \multirow{4}{*}{\rotatebox[origin=c]{30}{CWE}} & \multirow{4}{*}{\rotatebox[origin=c]{30}{FWE}} & \multirow{4}{*}{\rotatebox[origin=c]{30}{VT}} & \multirow{4}{*}{Avg.} \\
\cmidrule(lr){2-4}\cmidrule(lr){5-7}
& \rotatebox[origin=c]{30}{Single-1} & \rotatebox[origin=c]{30}{Single-2} & \rotatebox[origin=c]{30}{Single-3} & \rotatebox[origin=c]{30}{Multi-1} & \rotatebox[origin=c]{30}{Multi-2} & \rotatebox[origin=c]{30}{Multi-3} &  \\

\arrayrulecolor{black}\midrule

Vanilla   &99.67	&98.67	&100.0	&100.0	&100.0	&99.33	&93.25	&99.08	&97.80	&94.67	&100.0	&98.41\\
\arrayrulecolor{black!20}\midrule
\textit{FullReuse} &69.00	&78.67	&27.67	&23.67	&0.00	&1.33	&35.42	&38.08	&67.30	&92.78	&41.60	&43.23\\
\textit{PIE}   &98.00	&\textbf{92.67}	&94.00	&66.33	&82.67	&73.00	&65.00	&40.92	&86.10	&97.00	&50.93	&76.97\\
\textit{CacheBlend}   &\textbf{100.0}	&87.00	&96.00	&83.00	&83.89	&\textbf{78.67}	&\textbf{78.17}	&\textbf{57.75}	&\textbf{94.50}	&94.44	&69.27	&83.88\\
\textit{LegoLink}   &91.33	&83.33	&82.67	&78.00	&79.67	&74.33	&74.58	&46.17	&94.00	&97.00	&66.60	&78.88\\
\arrayrulecolor{black!20}\midrule
\textit{\textbf{ours}}   &98.00	&91.33	&\textbf{98.67}	&\textbf{83.67}	&\textbf{86.00}	&76.33	&75.92	&56.00	&92.57	&\textbf{97.11}	&\textbf{70.27}	&\textbf{84.17}\\

\arrayrulecolor{black}\bottomrule
\end{tabular}

\caption{Performance comparison of our method with \textit{FullReuse}, \textit{PIE}, \textit{CacheBlend}, \textit{LegoLink}, and up-bound method Vanilla on Ruler for Qwen2.5-7B-Inst. The best results are highlighted in \textbf{bold}.}
\label{table: ruler results}
\end{table*}

%% file: tables/ablations.tex
\begin{figure*}[h]
    \centering
    \begin{subfigure}[b]{0.33\textwidth}
        \includegraphics[width=\textwidth]{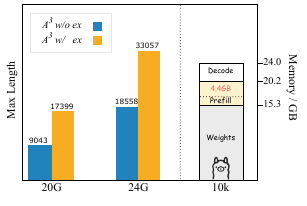}
        \caption{Longer generation and GPU memory savings with extension.}
        \label{fig:max_length}
    \end{subfigure}
    \hfill
    \begin{subfigure}[b]{0.33\textwidth}
        \hspace{-4mm} 
        \raisebox{-2mm}{ 
            \includegraphics[width=\textwidth]{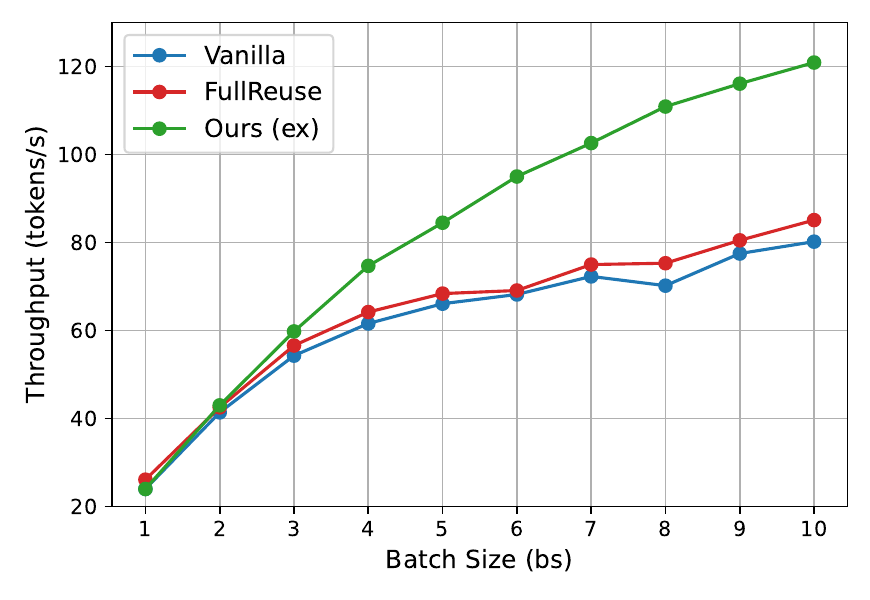}
        }
        \caption{Comparison of throughput between vanilla, the most efficient method, and $A^3$ (ex).}
        \label{fig:throughput}
    \end{subfigure}
    \hfill
    \begin{subfigure}[b]{0.31\textwidth}
        \raisebox{1mm}{ 
        \includegraphics[width=\textwidth]{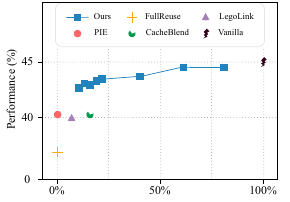}}
        \caption{Impact of recomputation ratio of our method on LongBench.}
        \label{fig:ratio}
    \end{subfigure}
    \caption{Inference performance variation of $A^3$ with extension and the ablation study on our method's hyperparameters.}
    \label{fig:all}
\end{figure*}

%% file: tables/ex.tex
\begin{table*}[t]

\centering
\setlength{\tabcolsep}{2.5pt} 
\begin{tabular}{l|c|ccccccccccc|c}

\specialrule{1pt}{0pt}{2pt}
\multicolumn{14}{c}{\textbf{LongBench}} \\ 
\midrule
 & Evict & \rotatebox[origin=c]{30}{Qasper} & \rotatebox[origin=c]{30}{MultiQ} & \rotatebox[origin=c]{30}{Hotpot} & \rotatebox[origin=c]{30}{2WikiM} & \rotatebox[origin=c]{30}{GovRep} & \rotatebox[origin=c]{30}{MultiN} & \rotatebox[origin=c]{30}{TREC} & \rotatebox[origin=c]{30}{TriviaQ} & \rotatebox[origin=c]{30}{SAMSum} & \rotatebox[origin=c]{30}{PCount} & \rotatebox[origin=c]{30}{LCC} & Avg. \\
\midrule
\multirow{2}{*}{\textbf{LLaMA}} 
& \ding{55} &39.07	&36.46	&42.17	&32.56	&28.88	&24.55	&68.67	&89.25	&39.33	&8.67	&65.39 &43.18 \\
& \ding{51} &36.19	&35.05	&42.31	&32.68	&25.84	&22.90	&67.00	&89.25	&37.64	&8.67	&61.34 &41.72 \\

\midrule
\multirow{2}{*}{\textbf{Mistral}} 
& \ding{55} &21.12	&39.31	&25.52	&18.38	&32.41	&25.28	&59.67	&76.44	&40.25	&3.15	&65.22	&36.98 \\
& \ding{51} &18.87	&39.42	&25.67	&17.28	&30.03	&24.70	&58.67	&82.65	&39.74	&2.67	&62.89	&36.60\\

\midrule
\multirow{2}{*}{\textbf{Qwen}} 
& \ding{55} &38.96	&41.18	&38.92	&25.77	&31.04	&22.61	&64.00	&86.56	&40.42	&4.87	&62.44	&41.52 \\
& \ding{51} &33.50	&39.38	&39.21	&26.04	&28.55	&21.55	&61.67	&85.98	&39.39	&6.03	&59.17 &40.04\\

\midrule
\multicolumn{14}{c}{\textbf{Ruler}} \\ 
\midrule

 & Evict & \rotatebox[origin=c]{30}{Single-1} & \rotatebox[origin=c]{30}{Single-2} & \rotatebox[origin=c]{30}{Single-3} & \rotatebox[origin=c]{30}{Multi-1} & \rotatebox[origin=c]{30}{Multi-2} & \rotatebox[origin=c]{30}{Multi-3} & \rotatebox[origin=c]{30}{MQuery} & \rotatebox[origin=c]{30}{MValue} & \rotatebox[origin=c]{30}{CWE} & \rotatebox[origin=c]{30}{FWE} & \rotatebox[origin=c]{30}{VT} & Avg. \\
 
\midrule
\multirow{2}{*}{\textbf{LLaMA}} 
& \ding{55}  &98.33	&97.33	&100.0	&79.00	&65.00	&71.00	&80.00	&65.50	&99.70	&98.22	&88.33	&85.67\\
& \ding{51} &98.33	&91.67	&3.33	&76.67	&48.67	&5.00	&78.17	&53.08	&97.10	&94.00	&88.67	&66.79 \\

\midrule
\multirow{2}{*}{\textbf{Mistral}} 
& \ding{55} &99.33	&0.33	&-	&0.33	&56.67	&-	&0.08	&0.50	&90.70	&70.56	&46.53	&33.18\\
& \ding{51} &99.33	&-	&-	&-	&16.67	&-	&11.08	&1.33	&85.63	&91.00	&49.13	&32.20 \\

\midrule
\multirow{2}{*}{\textbf{Qwen}} 
& \ding{55}  &98.00	&91.33	&98.67	&83.67	&86.00	&76.33	&75.92	&56.00	&92.57	&97.11	&70.27	&84.17 \\
& \ding{51} &98.00	&77.67	&12.00	&61.00	&4.33	&0.33	&37.25	&9.08	&82.63	&90.44	&71.00	&49.43 \\

\bottomrule
\end{tabular}

\caption{The impact of eviction-based extension on the task performance of our method. Experiments are conducted on three models and the LongBench, Ruler dataset.}
\label{table: ex}
\end{table*}

%% file: sections/Conclusion.tex
\section{Conclusion}
\label{sec:conclusion}

In this work, we propose $A^3$, an Attention-Aware Accurate KV Cache Fusion algorithm designed to address the performance degradation of existing KV reuse methods in long-context LLM inference. By leveraging query-aware attention to guide selective recomputation, $A^3$ effectively aligns KV updates with user-relevant information, mitigating attention loss without incurring significant overhead. Extensive experiments across multiple models and benchmarks demonstrate that $A^3$ outperforms prior reuse strategies in task performance while maintaining competitive decoding efficiency. Furthermore, we introduce an acceleration extension via token eviction, which complements $A^3$ by improving TPOT and overall throughput. Our findings highlight the feasibility and effectiveness of jointly optimizing KV reuse and eviction for practical long-context LLM applications.

%% file: sections/supplementary.tex
\appendix
\label{sec:appendix}

\section{Other Results}
\subsubsection{For LongBench.}
\begin{table*}[t]
\setlength{\tabcolsep}{4.1pt} 
\centering

\begin{tabular}{l|cccccccccccc}
\specialrule{1pt}{0pt}{2pt}
\multirow{4}{*}{Methods}  & \multicolumn{2}{c}{Single-Doc QA} & \multicolumn{2}{c}{Multi-Doc QA}& \multicolumn{2}{c}{Summarization}& \multicolumn{3}{c}{Few-shot Learning}& \multicolumn{2}{c}{Others} & \multirow{4}{*}{Avg.} \\
\cmidrule(lr){2-3}\cmidrule(lr){4-5}\cmidrule(lr){6-7}\cmidrule(lr){8-10}\cmidrule(lr){11-12}
& \rotatebox[origin=c]{30}{Qasper} & \rotatebox[origin=c]{30}{MultiQ} & \rotatebox[origin=c]{30}{Hotpot} & \rotatebox[origin=c]{30}{2WikiM} & \rotatebox[origin=c]{30}{GovRep} & \rotatebox[origin=c]{30}{MultiN} & \rotatebox[origin=c]{30}{TREC} & \rotatebox[origin=c]{30}{TriviaQ} & \rotatebox[origin=c]{30}{SAMSum} & \rotatebox[origin=c]{30}{PCount} & \rotatebox[origin=c]{30}{LCC}  & \\

\arrayrulecolor{black}\midrule
\multicolumn{13}{c}{Qwen2.5-7B-Instruct} \\
\arrayrulecolor{black!20}\midrule

Vanilla &40.10	&50.58	&53.76	&43.73	&33.70	&23.78	&67.00	&88.79	&42.82	&8.78	&62.99	&46.91\\
\arrayrulecolor{black!20}\midrule
\textit{FullReuse} &9.02	&23.82	&27.10	&16.83	&29.67	&22.09	&16.33	&65.40	&13.17	&2.81	&64.39	&26.42\\
\textit{PIE} &35.37	&40.06	&38.49	&23.74	&30.37	&22.19	&60.67	&84.25	&40.14	&4.29	&58.97	&39.87\\
\textit{CacheBlend} &37.07	&41.44	&\textbf{42.54}	&26.57	&31.29	&22.42	&61.33	&89.78	&40.37	&4.94	&60.38	&41.65\\
\textit{LegoLink} &37.67	&\textbf{42.72}	&40.98	&\textbf{28.70}	&\textbf{32.03}	&22.51	&60.67	&\textbf{89.79}	&\textbf{40.96}	&\textbf{5.47}	&62.13	&\textbf{42.15}\\
\textit{\textbf{ours}} &\textbf{38.96}	&41.18	&38.92	&25.77	&31.04	&\textbf{22.61}	&\textbf{64.00}	&86.56	&40.42	&4.87	&\textbf{62.44}	&41.52\\

\arrayrulecolor{black}\bottomrule
\end{tabular}

\caption{Performance comparison of our method with \textit{FullReuse}, \textit{PIE}, \textit{CacheBlend}, \textit{LegoLink}, and up-bound method Vanilla on LongBench for Qwen2.5-7B-Inst. The best results are highlighted in \textbf{bold}.}
\label{table: longbench results}
\end{table*}

We present the results of Qwen2.5-7B-Inst on LongBench. The results in Table \ref{table: longbench results} show that our method achieves comparable performance. Compared to selective recomputation methods such as \textit{CacheBlend} and $A^3$, \textit{LegoLink} even achieves better performance in some cases, which may be attributed to the more pronounced attention sink phenomenon in the Qwen model.

\subsubsection{For Needle-in-a-Haystack.} 
\begin{figure*}[t]
  \centering
  \textbf{(a) Qwen2.5-7B-Instruct Results} \\
  \includegraphics[width=0.33\linewidth]{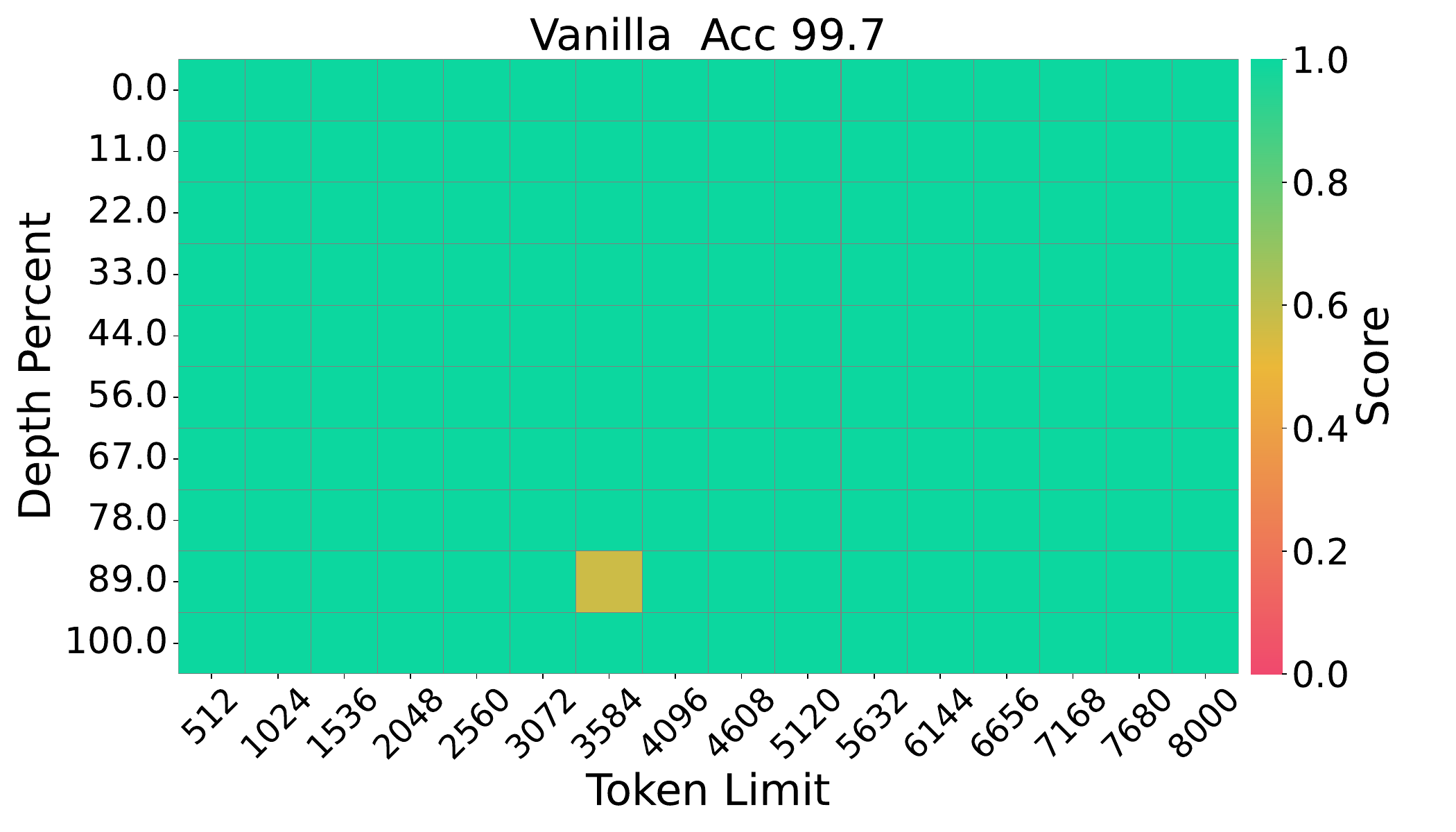}
  \includegraphics[width=0.33\linewidth]{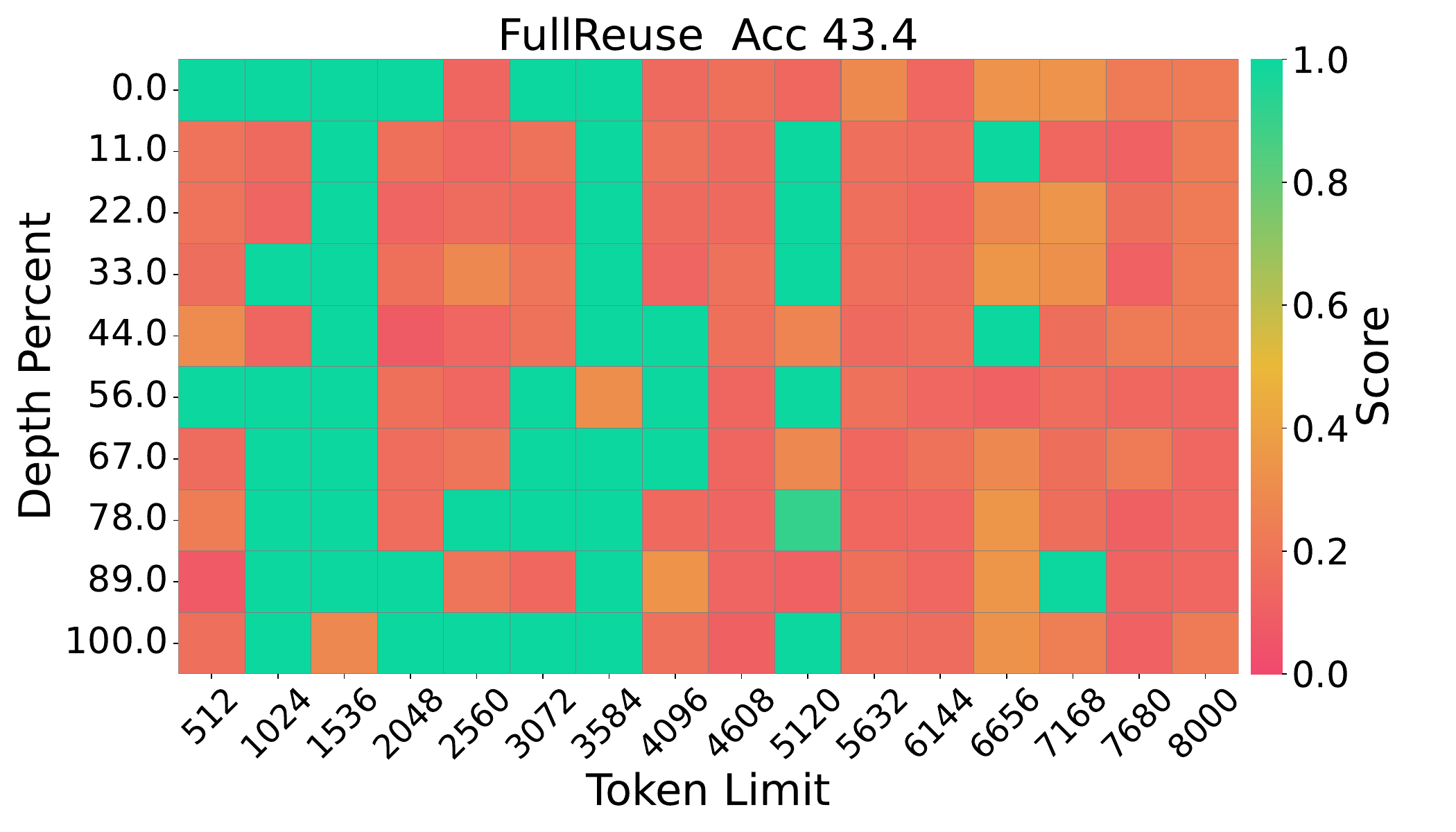}
  \includegraphics[width=0.33\linewidth]{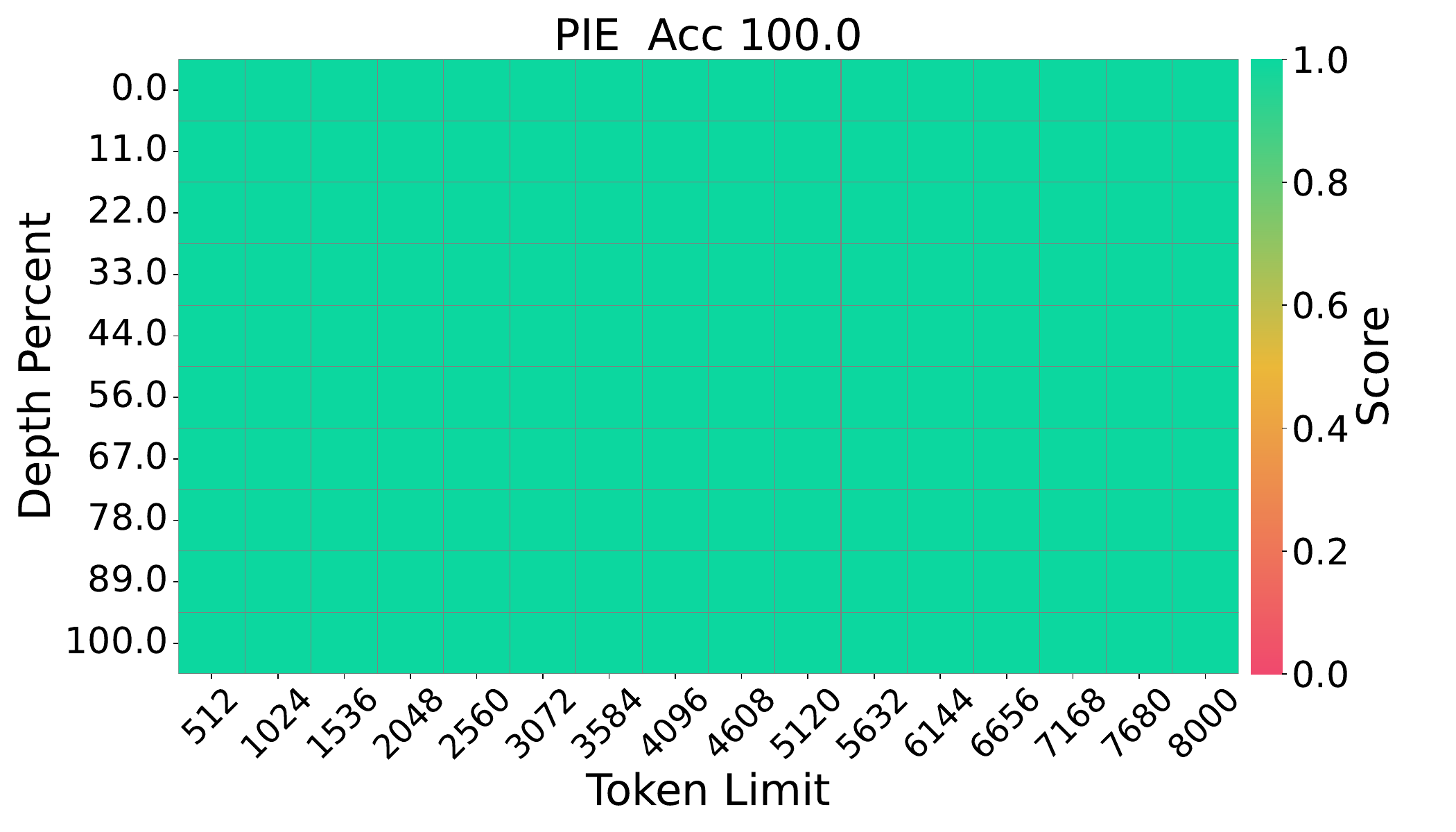} \\
  \includegraphics[width=0.33\linewidth]{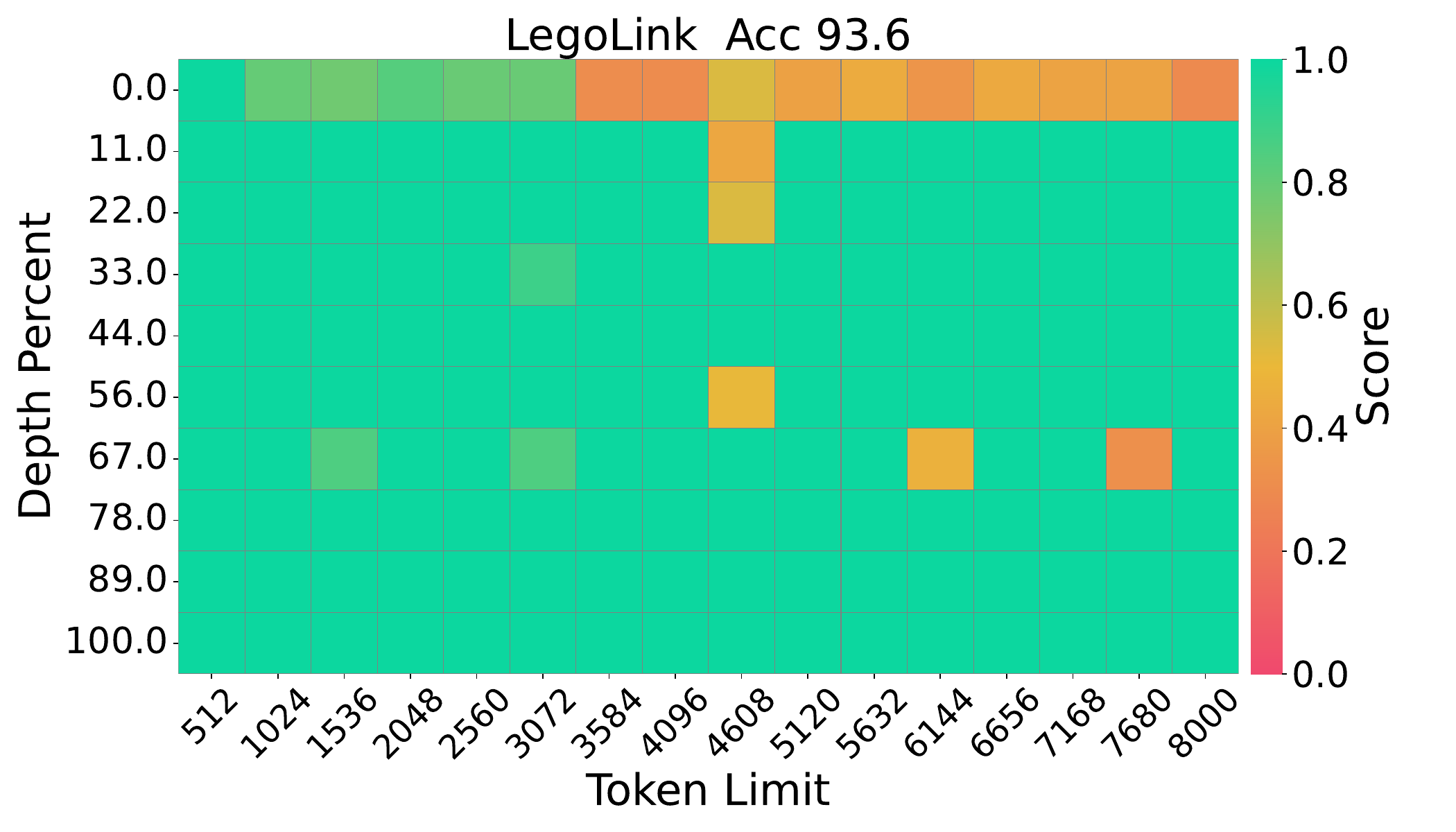}
  \includegraphics[width=0.33\linewidth]{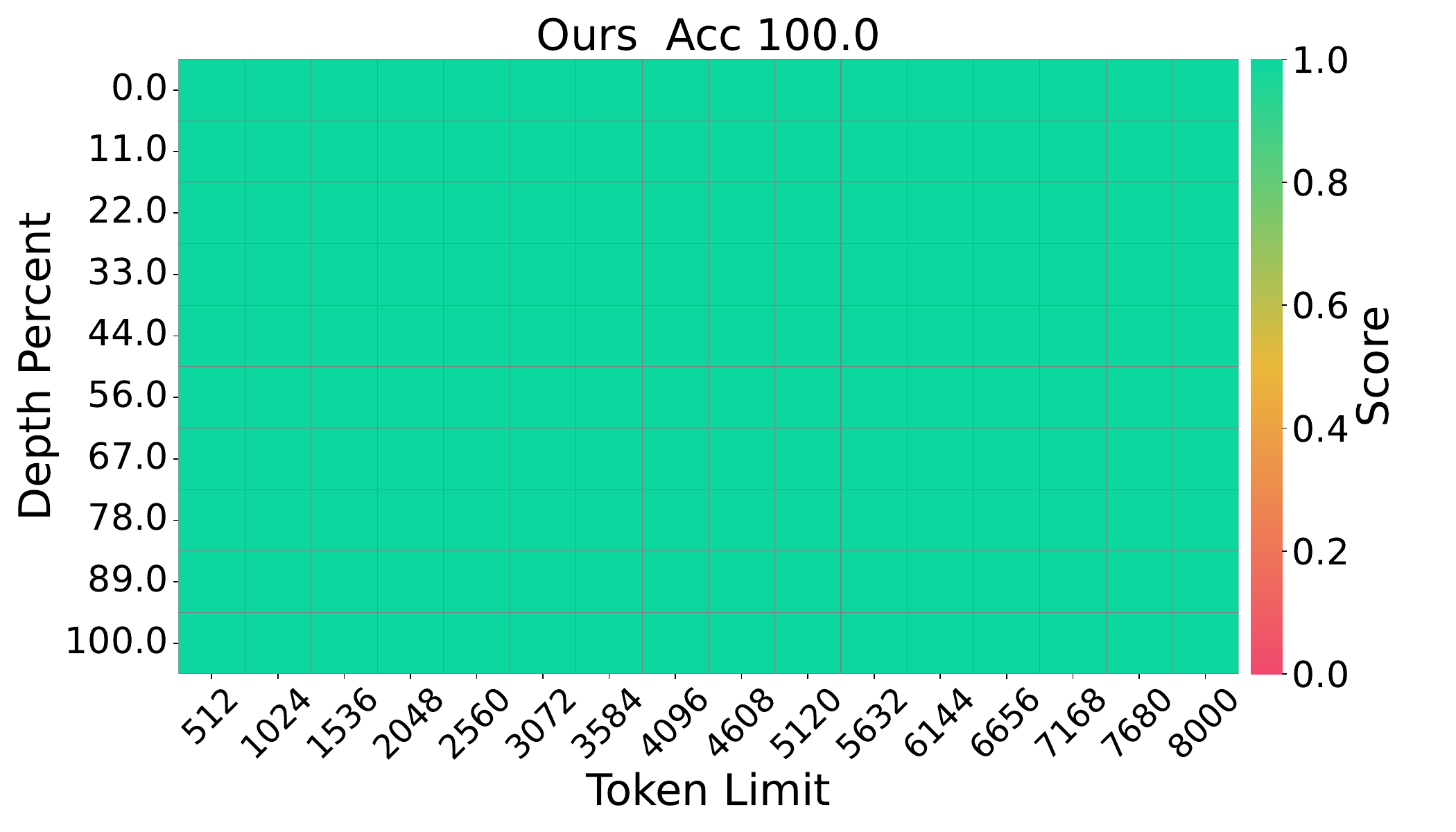}
  \includegraphics[width=0.33\linewidth]{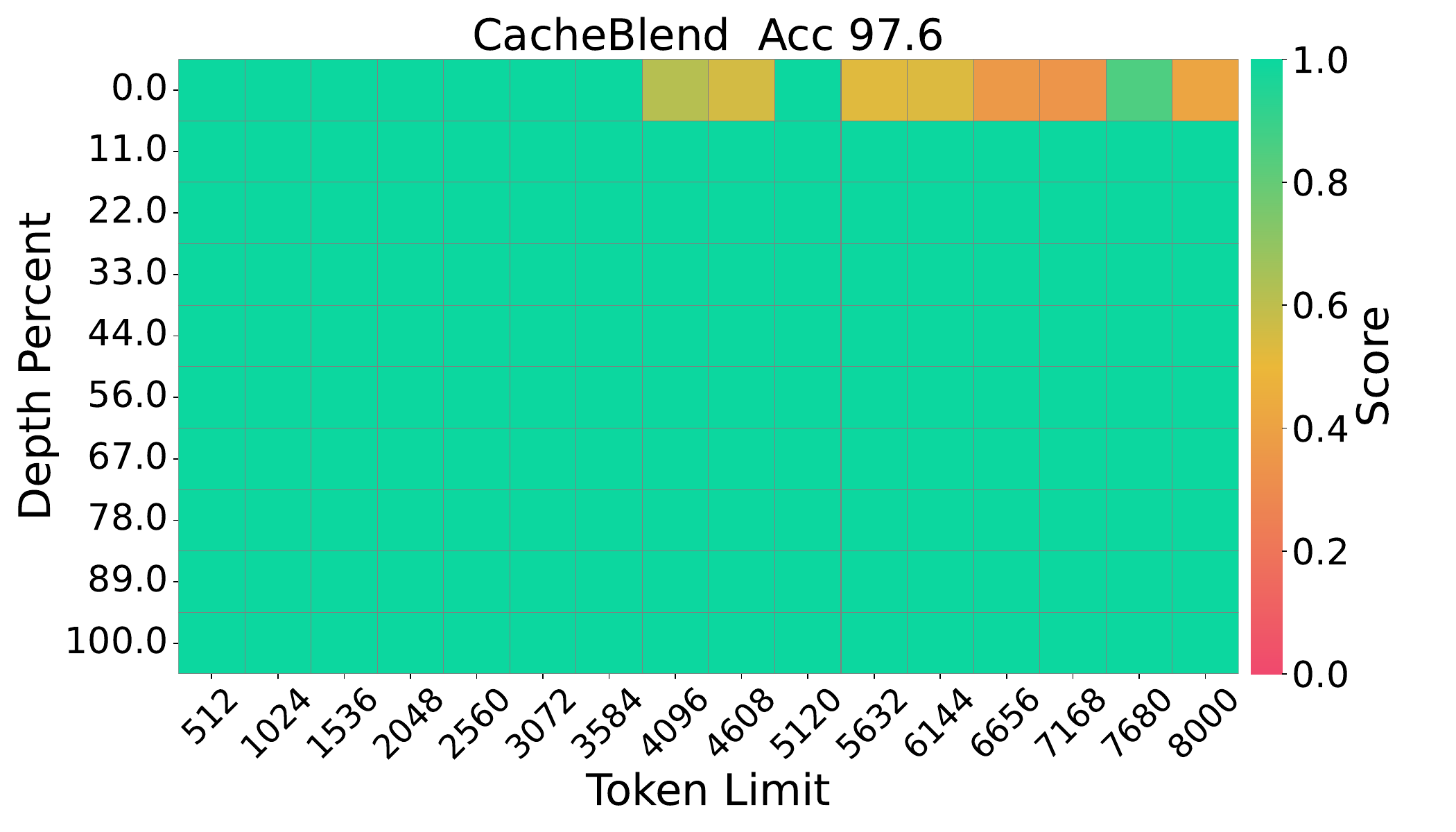} \\[1ex]

  \textbf{(b) Mistral-7B-Instruct-v0.2 Results} \\
  \includegraphics[width=0.33\linewidth]{figures/no.pdf}
  \includegraphics[width=0.33\linewidth]{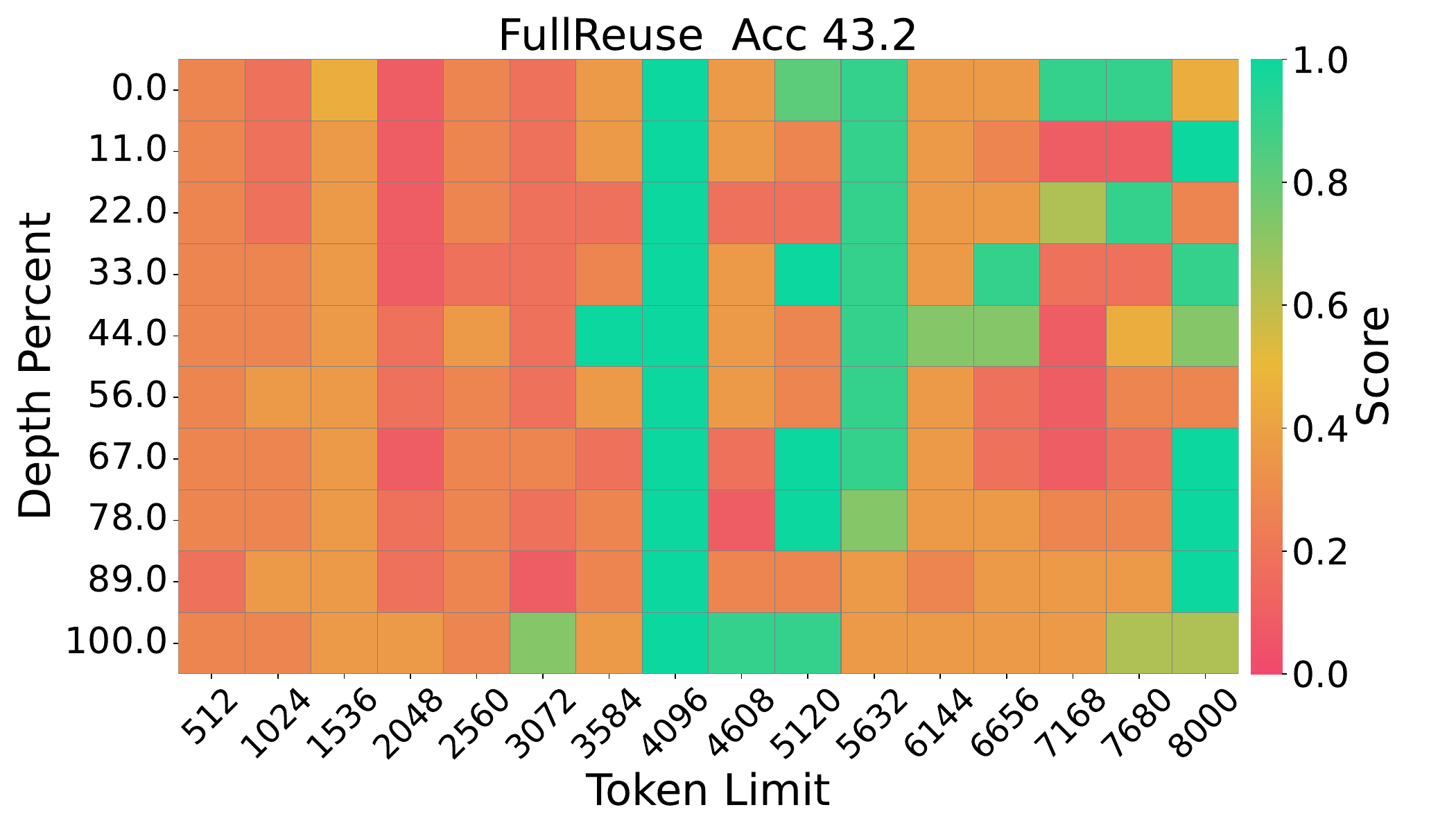}
  \includegraphics[width=0.33\linewidth]{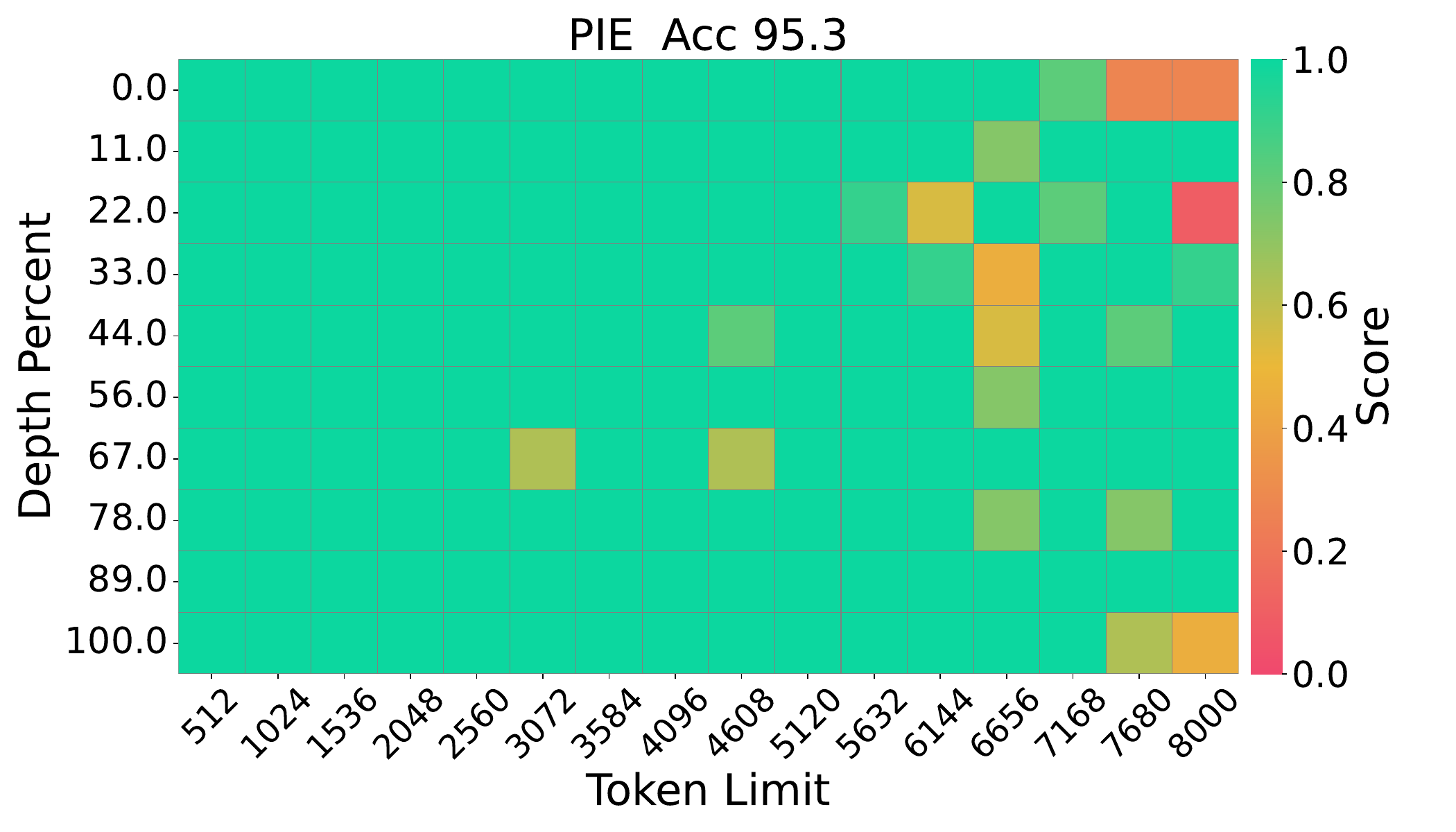} \\
  \includegraphics[width=0.33\linewidth]{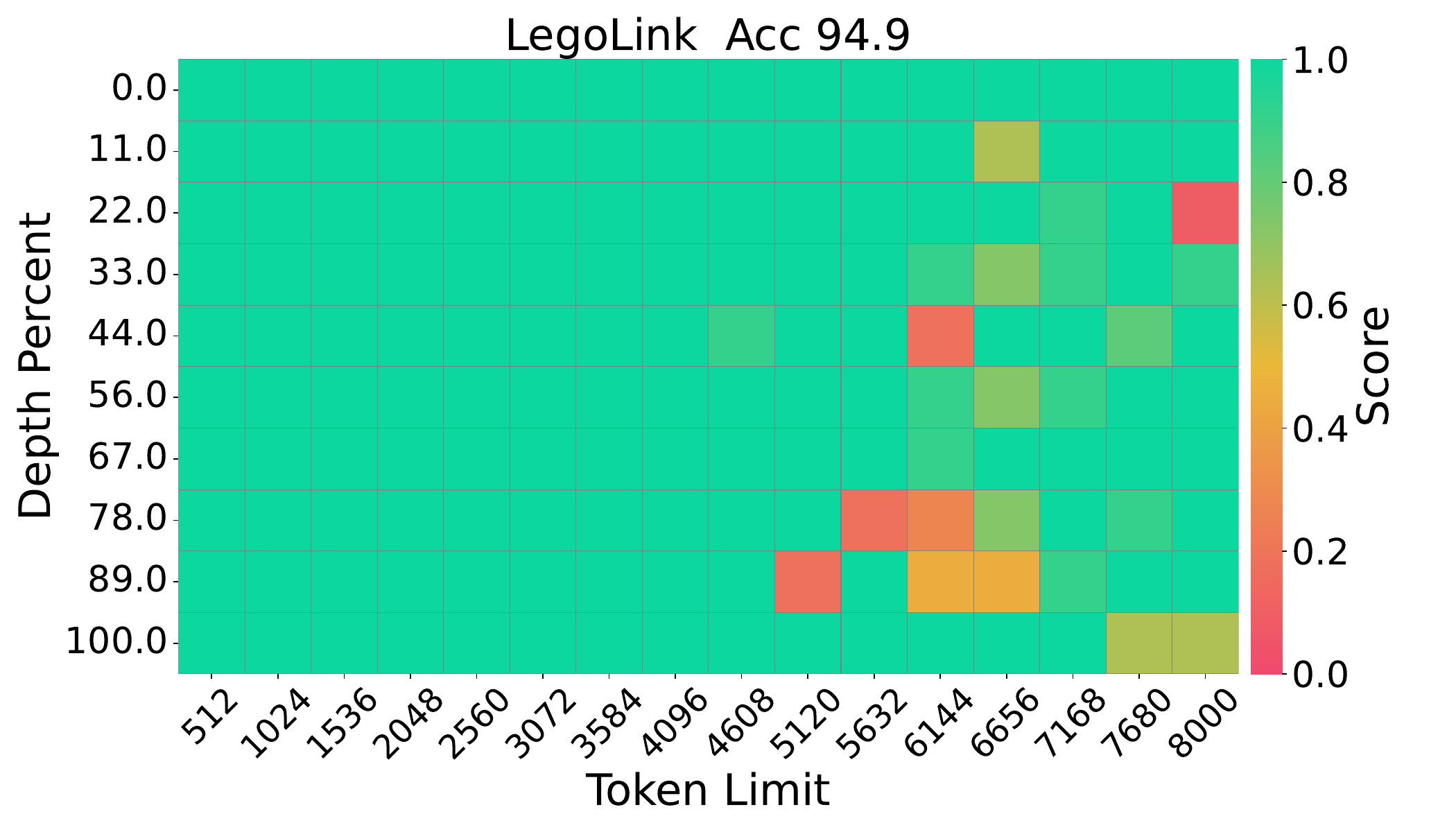}
  \includegraphics[width=0.33\linewidth]{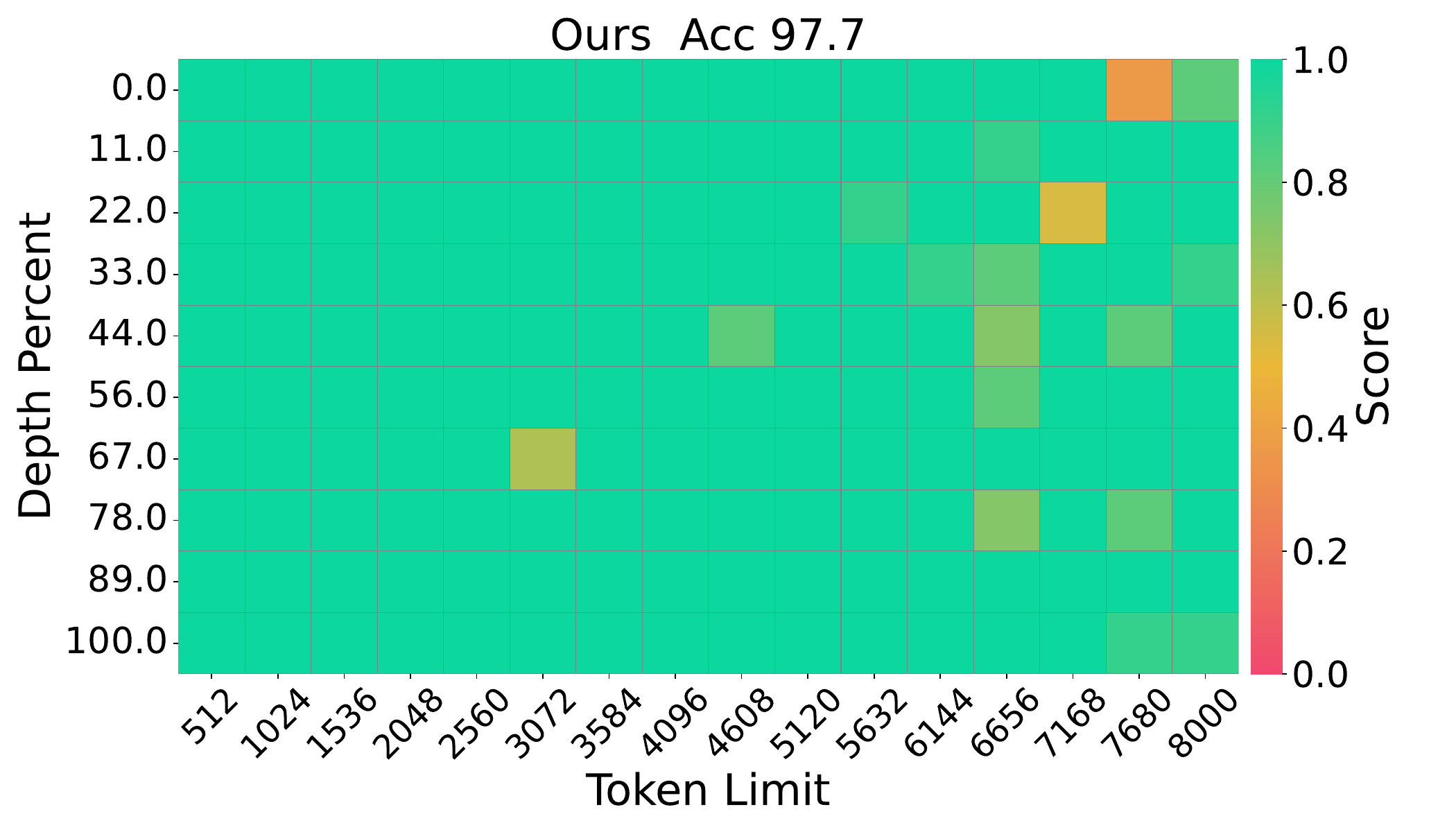}
  \includegraphics[width=0.33\linewidth]{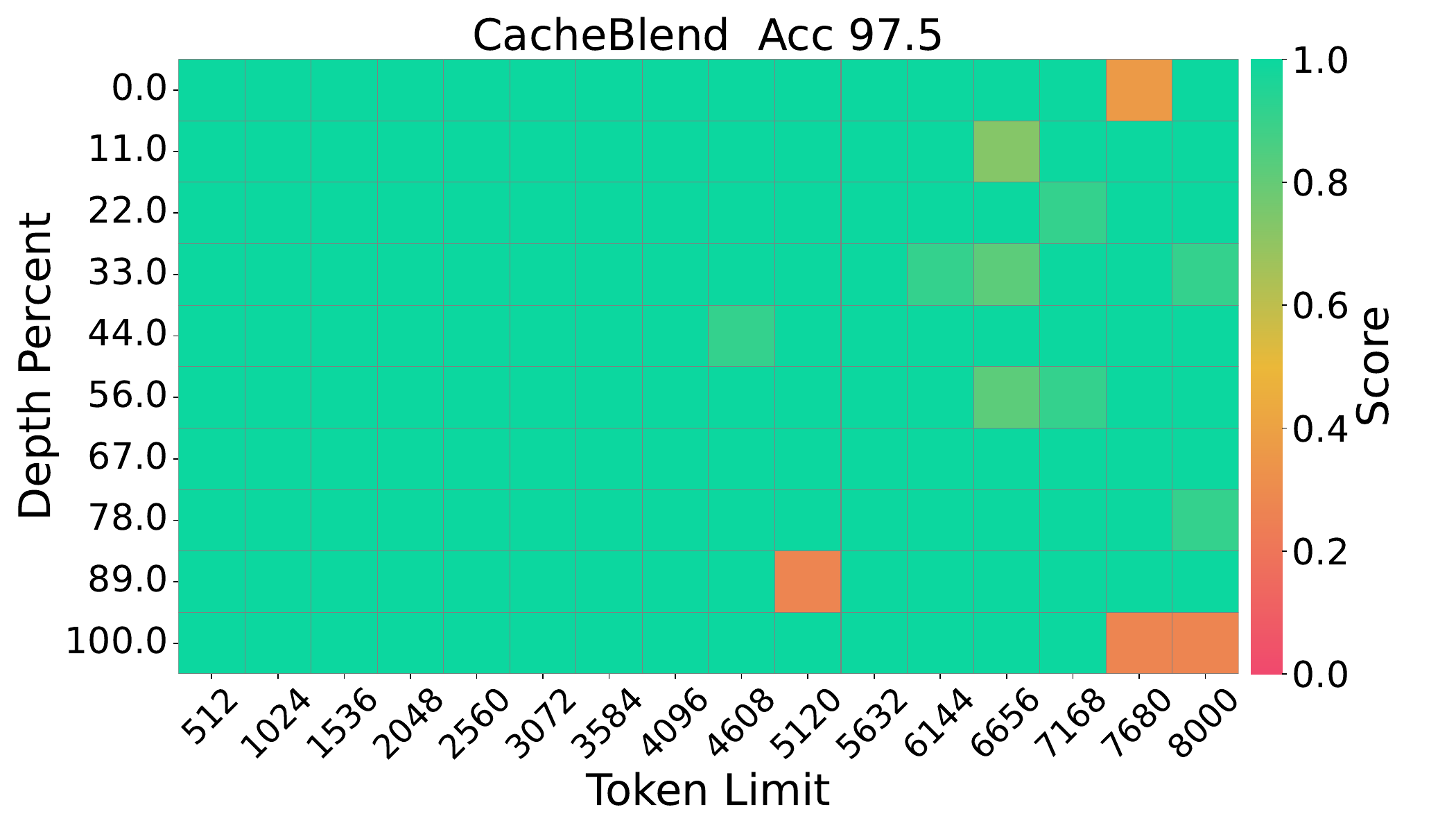}
  \caption{Needle-in-a-Haystack results on Qwen2.5-7B-Instruct and Mistral-7B-Instruct-v0.2. The vertical axis represents depth percentage, and the horizontal axis represents token length. Our method achieves the best retrieval performance.}
  \label{fig: needle-results}
\end{figure*}

We present the results on the Needle-in-a-Haystack task using Qwen2.5-7B-Instruct and Mistral-7B-Instruct-v0.2 in Figure \ref{fig: needle-results}. Our method consistently achieves the best performance. Notably, for the Qwen model, using position recovery alone is sufficient to achieve the same 100\% accuracy as our method.

\subsubsection{For Ruler.} We present the experimental results on the Ruler task using LLaMA3-8B-Instruct and Mistral-7B-Instruct-v0.2. Our method achieves the best average performance. In some configurations, the accuracy drops to zero, which may be attributed to limitations in model capability (e.g., zero in the Vanilla setting on Multi-3). However, all reuse methods also yield zero accuracy in these cases. In contrast, for other sub-datasets where valid results are observed, our method consistently demonstrates a strong ability to retrieve key information from long texts.

\section{Code Appendix}

\subsection*{Installation}
To set up the environment, please use the provided \texttt{requirements.txt} file:

\begin{verbatim}
pip install -r requirements.txt
\end{verbatim}

\subsection*{Usage}
All experiments are conducted under the \texttt{a3kv} directory.

\begin{enumerate}
    \item \textbf{Chunking Long Contexts}

    We provide scripts to chunk the datasets used in our evaluation:
    \begin{verbatim}
cd a3kv
python ./chunk_longbench.py
python ./chunk_needle.py
python ./chunk_ruler.py
    \end{verbatim}

    \item \textbf{Precomputation}

    The following script precomputes the attention cache for each chunked document:
    \begin{verbatim}
bash ./scripts/precompute.sh
    \end{verbatim}

    \item \textbf{Evaluation}

    We provide separate scripts to evaluate each benchmark:

    \begin{itemize}
        \item \textbf{LongBench:}
        \begin{verbatim}
bash ./scripts/eval_longbench.sh
        \end{verbatim}

        \item \textbf{Needle-in-a-Haystack:}
        \begin{verbatim}
bash ./scripts/eval_needle.sh
python ./visualize.py
        \end{verbatim}

        \item \textbf{Ruler:}
        \begin{verbatim}
bash ./scripts/eval_ruler.sh
python ./scripts/copy_results_ruler.py
        \end{verbatim}
    \end{itemize}
\end{enumerate}

\noindent

\begin{table*}[t]
\centering
\setlength{\tabcolsep}{4.2pt} 
\begin{tabular}{l|cccccccccccc}

\specialrule{1pt}{0pt}{2pt}
\multirow{4}{*}{Methods}  & \multicolumn{3}{c}{Single NIAH} & \multicolumn{3}{c}{Multi-key NIAH} & \multirow{4}{*}{\rotatebox[origin=c]{30}{MQuery}} & \multirow{4}{*}{\rotatebox[origin=c]{30}{MValue}} & \multirow{4}{*}{\rotatebox[origin=c]{30}{CWE}} & \multirow{4}{*}{\rotatebox[origin=c]{30}{FWE}} & \multirow{4}{*}{\rotatebox[origin=c]{30}{VT}} & \multirow{4}{*}{Avg.} \\
\cmidrule(lr){2-4}\cmidrule(lr){5-7}
& \rotatebox[origin=c]{30}{Single-1} & \rotatebox[origin=c]{30}{Single-2} & \rotatebox[origin=c]{30}{Single-3} & \rotatebox[origin=c]{30}{Multi-1} & \rotatebox[origin=c]{30}{Multi-2} & \rotatebox[origin=c]{30}{Multi-3} &  \\

\arrayrulecolor{black}\midrule
\multicolumn{13}{c}{LLaMA3-8B-Instruct} \\
\arrayrulecolor{black!20}\midrule

Vanilla  &100.0	&100.0	&100.0	&99.33	&99.67	&98.0	&91.33	&99.08	&99.73	&93.11	&99.53	&98.16 \\
\arrayrulecolor{black!20}\midrule
\textit{FullReuse}  &71.67	&83.67	&21.00	&29.33	&2.00	&0.00	&27.25	&41.50	&37.23	&96.33	&59.13 &42.65\\
\textit{PIE}  &94.33	&81.00	&99.00	&56.67	&55.00	&59.33	&57.08	&40.92	&99.40	&98.44	&64.80	&73.27 \\
\textit{CacheBlend}  &\textbf{100.0}	&97.00	&99.67	&78.33	&\textbf{67.00}	&69.67	&\textbf{81.17}	&58.67	&99.70	&98.22	&82.87	&84.75 \\
\textit{LegoLink}  &99.67	&90.00	&99.33	&71.67	&58.67	&65.67	&68.00	&54.75	&99.60	&\textbf{98.44}	&73.40	&79.93\\
\arrayrulecolor{black!20}\midrule
\textit{\textbf{ours}}  &98.33	&\textbf{97.33}	&\textbf{100.0}	&\textbf{79.00}	&65.00	&\textbf{71.00}	&80.00	&\textbf{65.50}	&\textbf{99.70}	&98.22	&\textbf{88.33}	&\textbf{85.67}\\

\arrayrulecolor{black}\midrule
\multicolumn{13}{c}{Mistral-7B-Instruct-v0.2} \\
\arrayrulecolor{black!20}\midrule

Vanilla   &94.67	&95.0	&63.67	&96.0	&8.33	&-	&83.50	&53.25	&67.93	&89.00	&85.60	&67.00\\
\arrayrulecolor{black!20}\midrule
\textit{FullReuse}  &97.33	&-	&-	&-	&0.67	&-	&-	&0.08	&87.90	&67.33	&43.33 &26.97\\
\textit{PIE}   &99.00	&-	&-	&0.33	&53.33	&-	&0.08	&0.42	&90.60	&71.56	&14.07	&29.94\\
\textit{CacheBlend}   &99.33	&-	&-	&-	&43.33	&-	&\textbf{0.17}	&0.17	&87.80	&\textbf{73.33}	&28.87	&30.27\\
\textit{LegoLink}   &99.33	&-	&-	&0.33	&51.00	&-	&-	&0.33	&89.93	&70.44	&30.93	&31.12\\
\arrayrulecolor{black!20}\midrule
\textit{\textbf{ours}}   &\textbf{99.33}	&\textbf{0.33}	&-	&\textbf{0.33}	&\textbf{56.67}	&-	&0.08	&\textbf{0.50}	&\textbf{90.70}	&70.56	&\textbf{46.53}	&\textbf{33.18}\\
\arrayrulecolor{black}\bottomrule
\end{tabular}

\caption{Performance comparison of our method with \textit{FullReuse}, \textit{PIE}, \textit{CacheBlend}, \textit{LegoLink}, and up-bound method Vanilla on Ruler for LLaMA3-8B-Instruct and Mistral-7B-Instruct-v0.2. The best results are highlighted in \textbf{bold}.}
\label{table: ruler results for others}
\end{table*}

\section{Limitations}

Although $A^3$ has achieved excellent results, there are still some limitations:
\begin{itemize}
    \item Our experiments are conducted exclusively on LLMs based on the Transformer architecture. Our method is not directly compatible with models built on alternative architectures or those that do not produce KV Caches, such as Mamba. Moreover, if a model does not use ROPE, position recovery during KV reuse becomes significantly more challenging.
    \item Additionally, KV reuse requires the offline storage (e.g., save on SSD) of a large number of KV Cache tensors. As a result, for large-scale external databases, the storage overhead may be substantial. Furthermore, loading precomputed KV Caches into the GPU imposes a heavy burden on the CPU, and high-concurrency requests may severely strain CPU performance.
\end{itemize}